\documentclass{article}

\usepackage{microtype}
\usepackage{graphicx}
\usepackage{subfigure}
\usepackage{booktabs} % for professional tables

\usepackage{hyperref}
\usepackage[accepted]{icml2024}

\usepackage{amsmath}
\usepackage{amssymb}
\usepackage{mathtools}
\usepackage{amsthm}

\usepackage{xspace}
\usepackage{graphicx}
\usepackage{float}
\usepackage{subfigure}
\usepackage{mathtools}
\usepackage{multirow}
\usepackage{makecell}
\usepackage{bm}
\usepackage{wrapfig}
\usepackage{bbding}
\usepackage{tablefootnote}
\usepackage{enumitem}
\usepackage{apxproof}
\usepackage{colortbl}
\usepackage[flushleft]{threeparttable}
\usepackage{natbib}
\usepackage{relsize}
\usepackage{capt-of}
\usepackage{tablefootnote}
\usepackage{siunitx}
\usepackage{hyperref} 
\hypersetup{
    linkcolor= red,                
}

\usepackage[capitalize,noabbrev]{cleveref}

\theoremstyle{plain}
\newtheorem{theorem}{Theorem}[section]
\newtheorem{proposition}[theorem]{Proposition}
\newtheorem{lemma}[theorem]{Lemma}

\theoremstyle{definition}
\newtheorem{definition}[theorem]{Definition}

\theoremstyle{remark}

\makeatletter
\DeclareRobustCommand\onedot{\futurelet\@let@token\@onedot}
\def\@onedot{\ifx\@let@token.\else.\null\fi\xspace}
 
\def\eg{\emph{e.g}\onedot} 
\def\ie{\emph{i.e}\onedot}

\def\aka{\emph{a.k.a}\onedot}

\makeatother

\usepackage[textsize=tiny]{todonotes}

\begin{document}

\twocolumn[
\icmltitle{Graph Distillation with Eigenbasis Matching}



\icmlsetsymbol{equal}{*}
% \icmlsetsymbol{corresponding}{$\dag$}

\begin{icmlauthorlist}
\icmlauthor{Yang Liu}{equal,yyy}
\icmlauthor{Deyu Bo}{equal,yyy}
\icmlauthor{Chuan Shi}{yyy}
\end{icmlauthorlist}

\icmlaffiliation{yyy}{Department of Computer Science, Beijing University of Posts and Telecommunication, Beijing, China}
\icmlcorrespondingauthor{Chuan Shi}{shichuan@bupt.edu.cn}

\icmlkeywords{Machine Learning, ICML}

\vskip 0.3in
]



\printAffiliationsAndNotice{\icmlEqualContribution} % otherwise use the standard text.

\begin{abstract}
\vspace{-3pt}
The increasing amount of graph data places requirements on the efficient training of graph neural networks (GNNs).
The emerging graph distillation (GD) tackles this challenge by distilling a small synthetic graph to replace the real large graph, ensuring GNNs trained on real and synthetic graphs exhibit comparable performance. 
However, existing methods rely on GNN-related information as supervision, including gradients, representations, and trajectories, which have two limitations.
First, GNNs can affect the spectrum (\ie, eigenvalues) of the real graph, causing \textit{spectrum bias} in the synthetic graph.
Second, the variety of GNN architectures leads to the creation of different synthetic graphs, requiring \textit{traversal} to obtain optimal performance.
% To tackle these issues, we propose Graph Distillation with Eigenbasis Matching (GDEM), which only aligns the eigenbasis and node features of real and synthetic graphs while directly replicating the spectrum of the real graph, thus preventing the influence of GNNs.
To tackle these issues, we propose Graph Distillation with Eigenbasis Matching (GDEM), which aligns the eigenbasis and
node features of real and synthetic graphs. Meanwhile, it
directly replicates the spectrum of the real graph
and thus prevents the influence of GNNs. Moreover, we design a discrimination constraint to balance the effectiveness and generalization of GDEM.
Theoretically, the synthetic graphs distilled by GDEM are restricted spectral approximations of the real graphs.
Extensive experiments demonstrate that GDEM outperforms state-of-the-art GD methods with powerful cross-architecture generalization ability and significant distillation efficiency.  Our code is available at \href{https://github.com/liuyang-tian/GDEM}{https://github.com/liuyang-tian/GDEM}.
\end{abstract}

\section{Introduction}
Graph neural networks (GNNs) are proven effective in a variety of graph-related tasks~\cite{GCN, GAT}.
However, the non-Euclidean nature of graph structure presents challenges to the efficiency and scalability of GNNs~\cite{SAGE}.
To accelerate training, one data-centric approach is to summarize the large-scale graph into a much smaller one.
Traditional methods primarily involve sparsification~\cite{RS, sparsification2} and coarsening~\cite{coarsening1, feature_coarsening}.
However, these methods are typically designed to optimize some heuristic metrics, \eg, spectral similarity~\cite{coarsening1} and pair-wise distance~\cite{spanner}, which may be irrelevant to downstream tasks, leading to sub-optimal performance.

Recently, graph distillation (GD), \aka, graph condensation, has attracted considerable attention in graph reduction due to its remarkable compression ratio and lossless performance~\cite{gd_survey}.
Generally, GD aims to synthesize a small graph wherein GNNs trained on it exhibit comparable performance to those trained on the real large graph.
To this end, existing methods are designed to optimize the synthetic graphs by matching some GNN-related information, such as gradients~\cite{GCond, DosCond}, representations~\cite{GCDM}, and training trajectories~\cite{SFGC}, between the real and synthetic graphs.
As a result, the synthetic graph aligns its distribution with the real graph and also incorporates information from downstream tasks.

\begin{figure}[t]
    \centering
    \subfigure{
        \centering
        \includegraphics[width=0.46\linewidth]{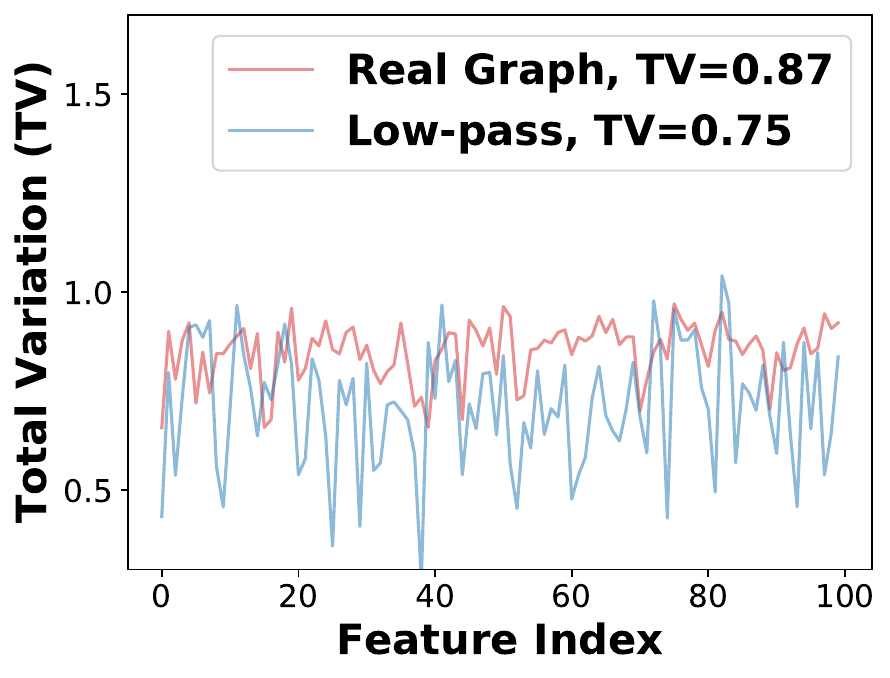}
    }
    \subfigure{
        \centering
        \includegraphics[width=0.46\linewidth]{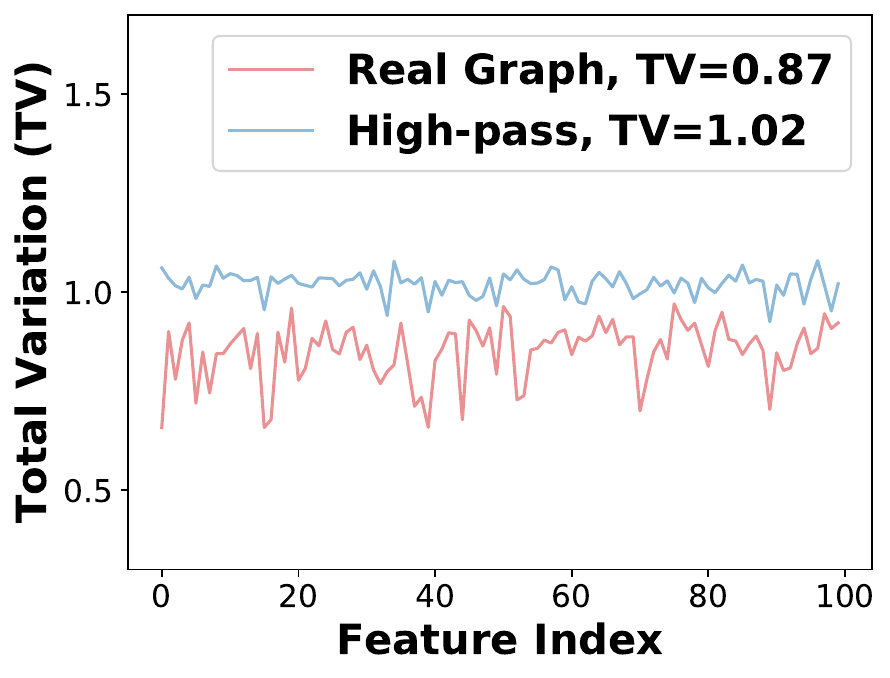}
    }
    \caption{Data distribution of the real and synthetic graphs in Pubmed dataset, where the average TV of the real graph is 0.87.
    \textbf{Left}: Synthetic graph distilled by a low-pass filter has a lower value of TV (0.75).
    \textbf{Right}: Synthetic graph distilled by a high-pass filter has a higher value of TV (1.02). For clarity, only the first 100-dimensional features are visualized. Best viewed in color.}
    \label{fig: TV_toy}
\end{figure}

Despite the considerable progress, existing GD methods require pre-selecting a specific GNN as the distillation model, introducing two limitations:
(1) GNNs used for distillation affect the real spectrum, leading to spectrum bias in the synthetic graph, \ie, a few eigenvalues dominate the data distribution.
Figure~\ref{fig: TV_toy} illustrates the total variation (TV)~\cite{TV} of the real and synthetic graphs.
Notably, TV reflects the smoothness of the signal over a graph. A small value of TV indicates a low-frequency distribution, and vice versa.
We can observe that the values of TV in the synthetic graph distilled by a low-pass filter consistently appear lower than those in the real graph, while the opposite holds for the high-pass filter, thus verifying the existence of spectrum bias (See Section~\ref{sec: upper-bound} for a theoretical analysis).
(2) The optimal performance is obtained by traversing various GNN architectures, resulting in non-negligible computational costs. Table~\ref{tab: pubmed_gcond} presents the cross-architecture results of GCOND~\cite{GCond} across six well-known GNNs, including GCN~\cite{GCN}, SGC~\cite{SGC}, PPNP~\cite{PPNP}, ChebyNet~\cite{ChebNet}, BernNet~\cite{BernNet}, and GPR-GNN~\cite{GPR-GNN}.
It can be seen that the evaluation performance of different GNNs varies greatly.
As a result, existing GD methods need to distill and traverse various GNN architectures to obtain optimal performance, which significantly improves the time overhead.
See Appendix~\ref{app: visualization} for the definition of TV and Appendix~\ref{app: cross} for more experimental details.

% These two observations state that GNN-related information has inherent bias and cannot be adapted to different architectures.

% We investigate the performance of GCOND~\cite{GCond} across six popular GNNs, namely:
% (1) Spatial methods, such as GCN~\cite{GCN}, SGC~\cite{SGC}, and PPNP~\cite{PPNP}.
% (2) Spectral methods, such as ChebyNet~\cite{ChebNet}, BernNet~\cite{BernNet}, and GPR-GNN~\cite{GPR-GNN}.
% The experiment proceeds as follows.
% First, six GNNs serve as the distillation model in turn, producing six different synthetic graphs.
% Subsequently, each synthetic graph is evaluated by six GNNs, resulting in $6\times6=36$ results, presented in Table~\ref{tab: pubmed_gcond}.
% Additionally, we visualize the total variation (TV)~\cite{TV} of the real and synthetic graphs in Figure~\ref{fig: TV_GCOND} to explore the distribution shift.
% We can find that
% (1) the values of TV in the synthetic graph distilled by GCOND are consistently lower than those in real graph;
% (2) the optimal performance of the six GNNs are obtained in different synthetic graphs.

\begin{table}[t]
    \centering
    \caption{Cross-architecture performance (\%) of GCOND with various distillation ($\mathtt{D}$) and evaluation ($\mathtt{E}$) GNNs in Pubmed dataset. \textbf{Bold} indicates the best in each column.}
    \resizebox{\linewidth}{!}{
    \begin{tabular}{lcccccc}
    \toprule
    $\mathtt{D}$ $\mathlarger{\diagdown}$ $\mathtt{E}$ & GCN & SGC & \multicolumn{1}{c}{PPNP} & Cheb. & Bern. & GPR. \\
    \midrule
    GCN   & 74.57 & 71.70 & 75.53 & 70.13 & 68.40 & 71.73 \\
    SGC   & \textbf{77.72} & \textbf{77.60} & 77.34 & 76.03 & 74.42 & 76.52 \\
    PPNP  & 72.70 & 70.40 & 77.46 & 73.38 & 70.56 & 74.02 \\
    Cheb. & 73.60 & 70.62 & 75.10 & \textbf{77.30} & 77.62 & 78.10 \\
    Bern. & 67.68 & 73.76 & 74.30 & 77.20 & \textbf{78.12} & \textbf{78.28} \\
    GPR.  & 76.04 & 72.20 & \textbf{77.94} & 75.92 & 77.12 & 77.96 \\
    \bottomrule
    \end{tabular}}
    \label{tab: pubmed_gcond}
\end{table}

Once the weaknesses of existing methods are identified, it is natural to ask: \textit{How to distill graphs without being affected by different GNNs?}
% To answer this question, we initially derive an upper bound on the objective of gradient matching (Section~\ref{sec: upper-bound}) and find that the data distribution in the real graph is dominated by the principal eigenvalues in the filtering function of GNNs.
To answer this question, we propose Graph Distillation with Eigenbasis Matching (GDEM).
% to mitigate the influence of GNNs.
Specifically, GDEM decomposes the graph structure into eigenvalues and eigenbasis.
During distillation, GDEM matches the eigenbasis and node features of real and synthetic graphs, which equally preserves the information of different frequencies, thus addressing the spectrum bias.
Additionally, a discrimination loss is jointly optimized to improve the performance of GDEM and balance its effectiveness and generalization.
Upon completing the matching, GDEM leverages the real graph spectrum and synthetic eigenbasis to construct a complete synthetic graph, which prevents the spectrum from being affected by GNNs and ensures the uniqueness of the synthetic graph, thus avoiding the traversal requirement and improving the distillation efficiency.

The contributions of our paper are as follows.
(1) We systematically analyze the limitations of existing distillation methods, including spectrum bias and traversal requirement.
(2) We propose GDEM, a novel graph distillation framework, which mitigates the dependence on GNNs by matching the eigenbasis instead of the entire graph structure. Additionally, it is theoretically demonstrated that GDEM preserves essential spectral similarity during distillation.
(3) Extensive experiments on seven graph datasets validate the superiority of GDEM over state-of-the-art GD methods in terms of effectiveness, generalization, and efficiency.

% \begin{table*}[t]
%     \begin{minipage}{0.37\linewidth}
%         \centering
%         \caption{Time overhead of GCOND and GDEM.}
%         \resizebox{\linewidth}{!}{
%         \begin{tabular}{lccc}
%         \toprule
%         & \makecell{GCOND\\(SGC)} & \makecell{GCOND\\(Traversal)} & GDEM \\
%         \midrule
%         Time ($s$)   & 76.1 & 74.4 & 78.0 \\
%         \bottomrule
%         \end{tabular}}
%         \hfill
%         \vspace{-10pt}
%         \centering
%         \caption{Optimal results of GCOND and GDEM.}
%         \resizebox{\linewidth}{!}{
%         \begin{tabular}{lcccccc}
%         \toprule
%         & GCN & SGC & \multicolumn{1}{c}{PPNP} & Cheb. & Bern. & GPR. \\
%         \midrule
%         GCOND   & 76.1 & 74.4 & 78.0 & 77.3 & \textbf{78.2} & 78.3 \\
%         GDEM    & \textbf{78.4} & \textbf{76.1} & \textbf{78.1} & \textbf{78.1} & \textbf{78.2} & \textbf{78.6} \\
%         \bottomrule
%         \end{tabular}}
%     \end{minipage}%
% \end{table*}

\section{Preliminary}
Before describing our framework in detail, we first introduce some notations and concepts used in this paper.
Specifically, we focus on the node classification task, where the goal is to predict the labels of the nodes in a graph.
Assume that there is a graph $\mathcal{G}=(\mathcal{V}, \mathcal{E}, \mathbf{X})$, where $\mathcal{V}$ is the set of nodes with $|\mathcal{V}|=N$, $\mathcal{E}$ indicates the set of edges, and $\mathbf{X} \in \mathbb{R}^{N \times d}$ is the node feature matrix.
The adjacency matrix of $\mathcal{G}$ is defined as $\mathbf{A} \in \{0, 1\}^{N \times N}$, where $A_{ij}=1$ if there is an edge between nodes $i$ and $j$, and $A_{ij} =0$ otherwise. The corresponding normalized Laplacian matrix is defined as $\mathbf{L} = \mathbf{I}_{N} - \mathbf{D}^{-\frac{1}{2}} \mathbf{A} \mathbf{D}^{-\frac{1}{2}}$, where $\mathbf{I}_{N}$ is an identity matrix and $\mathbf{D}$ is the degree matrix with $D_{ii}=\sum_{j} A_{ij}$ for $i \in \mathcal{V}$ and $D_{ij} = 0$ for $i \neq j$.
Without loss of generality, we assume that $\mathcal{G}$ is undirected and all the nodes are connected.

\vspace{-5pt}
\paragraph{Eigenbasis and Eigenvalue.}
The normalized graph Laplacian can be decomposed as $\mathbf{L} = \mathbf{U} \bm{\Lambda} \mathbf{U}^{\top} = \sum_{i=1}^{N}\lambda_{i}\mathbf{u}_{i}\mathbf{u}_{i}^{\top}$, where $\bm{\Lambda}=\text{diag}(\{\lambda_{i}\}_{i=1}^{N})$ are the eigenvalues and $\mathbf{U}=[\mathbf{u}_{1}, \cdots, \mathbf{u}_{N}] \in \mathbb{R}^{N \times N}$ is the eigenbasis, consisting of a set of eigenvectors.
Each eigenvector $\mathbf{u}_{i} \in \mathbb{R}^{N}$ has a corresponding eigenvalue $\lambda_{i}$, such that $\mathbf{L}\mathbf{u}_{i} = \lambda_{i}\mathbf{u}_{i}$.
Without loss of generality, we assume $0 \leq \lambda_{1} \leq \cdots \leq \lambda_{N} \leq 2$.

% The eigenbasis is usually used as the basis of graph Fourier transform and its inverse~\cite{GSP}, which offers a way to define the graph convolution operator:
% \begin{equation}
%     \mathbf{x} *_G f = \mathbf{U} \left( \left(\mathbf{U}^{\top} f \right) \odot \left(\mathbf{U}^{\top} \mathbf{x} \right) \right),
% \end{equation}
% where $\mathbf{x} \in \mathbb{R}^{n \times 1}$ is a graph signal and $f$ is a spatial filter.
% According to the convolution theorem, convolution in the spatial domain $*_G$ is equal to the multiplication $\odot$ in the spectral domain.

% Replacing the spatial filter $\mathbf{U}^{\top} f$ with a learnable diagonal matrix $\mathbf{G}=\text{diag}(\{g_{i}\}_{i=1}^{n})$ yields the architecture of modern GNNs:
% \begin{equation}
%     \mathbf{x}^{\prime} = \mathbf{U} \mathbf{G} \mathbf{U}^{\top} \mathbf{x} = \sum_{i=1}^{n} g_{i} \mathbf{u}_{i} \mathbf{u}_{i}^{\top} \mathbf{x},
% \end{equation}
% where $\mathbf{x}^{\prime}$ is the filtered graph signal.
% Note that the graph filter $\mathbf{U} \mathbf{G} \mathbf{U}^{\top}$ forms a new graph structure that is a weighted sum of the subspaces of eigenbasis, \eg, $\mathbf{U} \mathbf{G} \mathbf{U}^{\top} = \sum_{i=1}^{n} g_{i} \mathbf{u}_{i} \mathbf{u}_{i}^{\top}$.

\vspace{-5pt}
\paragraph{Graph Distillation.}
GD aims to distill a small synthetic graph $\mathcal{G}^{\prime}=(\mathcal{V}^{\prime}, \mathcal{E}^{\prime}, \mathbf{X}^{\prime})$, where $|\mathcal{V}^{\prime}| = N^{\prime} \ll N$ and $\mathbf{X}^{\prime} \in \mathbb{R}^{N^\prime \times d}$,
from the real large graph $\mathcal{G}$.
Meanwhile, GNNs trained on $\mathcal{G}$ and $\mathcal{G}^{\prime}$ will have comparable performance, thus accelerating the training of GNNs.
Existing frameworks can be divided into three categories: gradient matching, distribution matching, and trajectory matching.
See Appendix~\ref{app: gd} for more detailed descriptions.

\section{Spectrum Bias in Gradient Matching}
\label{sec: upper-bound}

In this section, we give a detailed analysis of the objective of gradient matching in graph data, which motivates the design of our method.
We start with a vanilla example, which adopts a one-layer GCN as the distillation model and simplifies the objective of GNNs into the MSE loss:
\begin{equation}
    \mathcal{L} = \frac{1}{2} \left\| \mathbf{A} \mathbf{X} \mathbf{W} - \mathbf{Y} \right\|_{F}^{2},
\end{equation}

\vspace{-10pt}
where $\mathbf{W}$ is the model parameter.
The gradients on the real and synthetic graphs are calculated as follows:
\vspace{-5pt}
\begin{equation}
\begin{aligned}
    \nabla_{\mathbf{W}} &= { \left( \mathbf{A} \mathbf{X} \right)}^T \left( \mathbf{A} \mathbf{X} \mathbf{W} - \mathbf{Y} \right), \\
    \nabla_{\mathbf{W}}^{\prime} &= { \left( \mathbf{A}^{\prime} \mathbf{X}^{\prime} \right)}^T \left( \mathbf{A^{\prime}} \mathbf{X}^{\prime} \mathbf{W} - \mathbf{Y}^{\prime} \right).
\end{aligned}
\end{equation}

\vspace{-10pt}
Assume that the objective of gradient matching is the MSE loss between two gradients, \ie, $\mathcal{L}_{GM} = \| \nabla_{\mathbf{W}} - \nabla_{\mathbf{W}}^{\prime} \|_{F}^{2}$.
To further characterize its properties, we analyze the following upper-bound of $\mathcal{L}_{GM}$:
\vspace{-5pt}
\begin{equation}
\begin{aligned}
    \mathcal{L}_{GM} \leq
    &\| \mathbf{W} \|^{2}_{F} \| \mathbf{X}^{\top}\mathbf{A}^{2}\mathbf{X} - {\mathbf{X}^{\prime}}^{\top} {\mathbf{A}^{\prime}}^{2} \mathbf{X}^{\prime} \|^{2}_{F} \\
    &+ \| \mathbf{X}^{\top}\mathbf{A}\mathbf{Y} - {\mathbf{X}^{\prime}}^{\top} \mathbf{A}^{\prime} \mathbf{Y}^{\prime} \|^{2}_{F},
\label{eq: L_GM}
\end{aligned}
\end{equation}
where $\mathbf{X}^{\top}\mathbf{A}^{2}\mathbf{X}$ and $\mathbf{X}^{\top}\mathbf{A}\mathbf{Y}$ are two target distributions in the real graph, which are used to supervise the update of the synthetic graph.
However, both of them will be dominated by a few eigenvalues, resulting in spectrum bias.

\begin{lemma}
    The target distribution of GCN is dominated by the smallest eigenvalue after stacking multiple layers.
\label{collpse_GCN}
\end{lemma}
\vspace{-15pt}
\begin{proof}
    The target distribution can be reformulated as:
    \vspace{-5pt}
    \begin{equation}
        \mathbf{X}^{\top}\mathbf{A}^{2t}\mathbf{X} = \sum_{i=1}^{N} (1 - \lambda_{i})^{2t} \mathbf{X}^{\top} \mathbf{u}_{i} \mathbf{u}_{i}^{\top} \mathbf{X},
    \end{equation}
    where $t$ is the number of layers.
    When $t$ goes to infinity, only the smallest eigenvalue $\lambda_{0}=0$ preserves its coefficient $(1 - \lambda_{0})^{2t}=1$ and other coefficients tend to 0.
    Hence, the target distribution $\mathbf{X}^{\top}\mathbf{A}^{2t}\mathbf{X}$ is dominated by $\mathbf{X}^{\top}\mathbf{u}_{0}\mathbf{u}_{0}^{\top}\mathbf{X}$.
    The same analysis can be applied for $\mathbf{X}^{\top}\mathbf{A}^{t}\mathbf{Y}$.
\end{proof}

\begin{lemma}
    Suppose the distillation GNN has an analytic filtering function $g(\cdot)$. Then the target distributions will be dominated by the eigenvalues whose filtered values are greater than 1, \ie, $g(\lambda_{i}) \geq 1$.
\label{collapse_GNN}
\end{lemma}
\vspace{-15pt}
\begin{proof}
    The objective function of distillation GNN is
    $\mathcal{L} = \frac{1}{2} \left\| g(\mathbf{L}) \mathbf{X} \mathbf{W} - \mathbf{Y} \right\|_{F}^{2}$.
    Then the target distributions become $\mathbf{X}^{\top}g(\mathbf{L})^{2t}\mathbf{X}$ and $\mathbf{X}^{\top}g(\mathbf{L})^{t}\mathbf{Y}$ as $g$ is analytic.
    Therefore, the filtered eigenvalues with values $g(\lambda_{i}) \geq 1$ retain their coefficients and dominate the target distributions.
\end{proof}
Lemmas~\ref{collpse_GCN} and~\ref{collapse_GNN} state that leveraging the information of GNNs in distillation will introduce a spectral bias in the target distributions.
As a result, the synthetic graph can only match part of the data distribution of the real graph, leaving its structural information incomplete.

% Replacing the one-layer GCN with other graph filters results in different GNNs.
% For example, PPNP~\cite{PPNP} defines the filter as $g_{i} = \alpha + (1 - \alpha)\lambda_{i}$, where $\alpha$ is a hyperparameter.
% This replacement does not affect our analysis but provides different spectrum preferences.
% On the other hand, despite the simplifications in our analysis, certain issues remain unaddressed.
% For instance, the impact of spectrum bias on the structure of synthetic graphs.
% We leave them for future investigations.

% \begin{remark}
%     (Motivation) In graph condensation, the usage of graph filters or GNNs biases the synthetic graph toward a specific spectrum, which cannot preserve the entire distribution of the real graph.
%     To address this problem, it is important to avoid using the spectrum information during the condensation process.
% \end{remark}

\section{The Proposed Method: GDEM}

\begin{figure}[t]
    \centering
    \subfigure[Gradient matching]{
        \centering
        \includegraphics[width=0.46\linewidth]{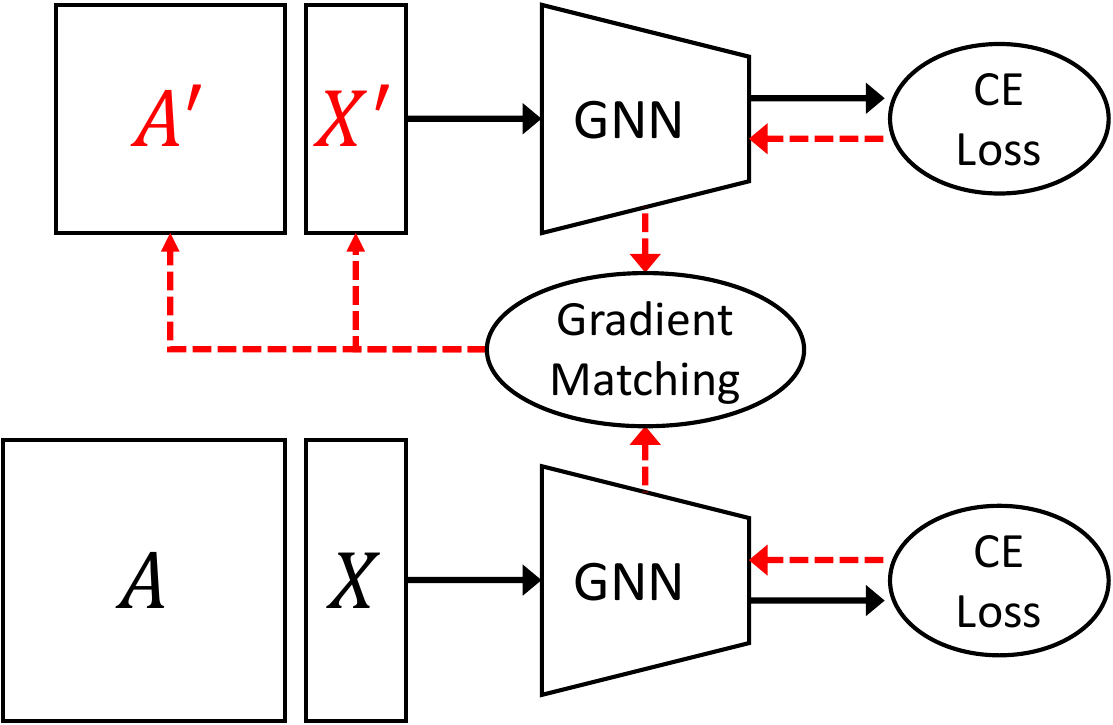}
        \label{fig: GM}
    }
    \subfigure[Distribution matching]{
        \centering
        \includegraphics[width=0.46\linewidth]{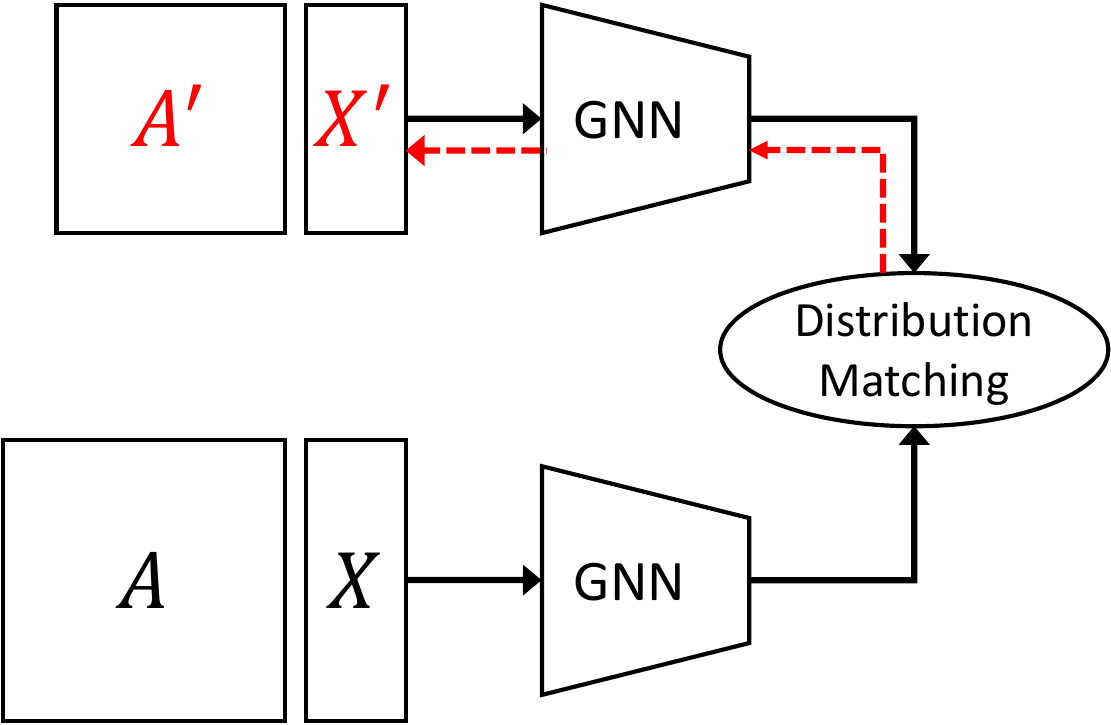}
        \label{fig: DM}
    }
    \quad
    \subfigure[Eigenbasis matching]{
        \includegraphics[width=\linewidth]{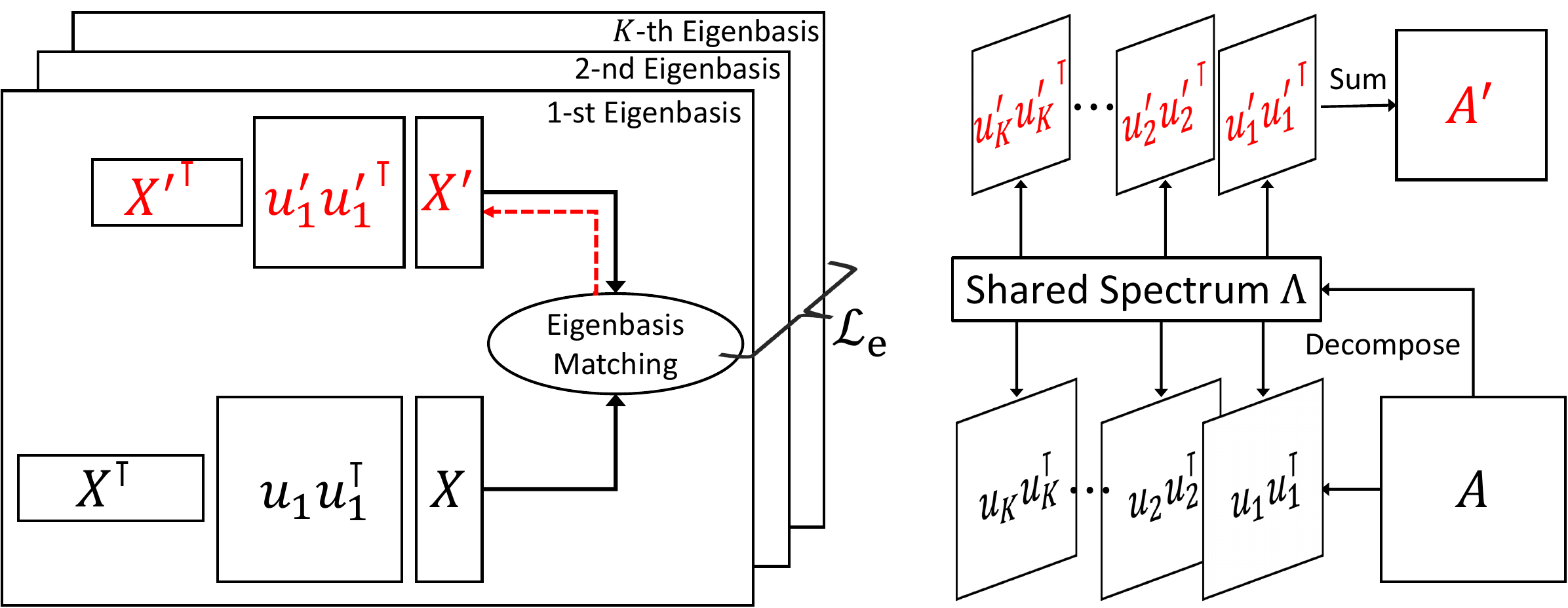}
        \label{fig: EM}
    }
    \caption{Comparison between different graph distillation methods, where the red characters represent the synthetic data, the solid black lines, and red dotted lines indicate the forward and backward passes, respectively.}
    % (a) Gradient matching matches the model gradients of Cross-Entropy (CE) loss.
    % (b) Distribution matching matches the node representations learned by the distillation GNNs.
    % (c) Eigenbasis matching first matches the node features in the subspaces defined by the eigenbasis, and then constructs the adjacency matrix by leveraging the spectrum of the real graph.}
\end{figure}

In this section, we introduce the proposed method GDEM. 
Compared with previous methods, \eg, gradient matching (Figure~\ref{fig: GM}) and distribution matching (Figure~\ref{fig: DM}), GDEM, illustrated in~\ref{fig: EM}, does not rely on specific GNNs, whose distillation process can be divided into two steps:
(1) Matching the eigenbasis and node features between the real and synthetic graphs.
(2) Constructing the synthetic graph by using the synthesized eigenbasis and real spectrum.

\subsection{Eigenbasis Matching}

The eigenbasis of a graph represents its crucial structural information. For example, eigenvectors corresponding to smaller eigenvalues reflect the global community structure, while eigenvectors corresponding to larger eigenvalues encode local details~\cite{FAGCN}.
Generally, the number of eigenvectors is the same as the number of nodes in a graph, suggesting that we cannot preserve all the real eigenbasis in the synthetic graph.
Therefore, GDEM is designed to match eigenvectors with the $K_{1}$ smallest and the $K_{2}$ largest eigenvalues, where $K_{1}$ and $K_{2}$ are hyperparameters, and $K_1 + K_2 = K \leq N^{\prime}$.
This approach has been proven effective in both graph coarsening~\cite{k1+k2} and spectral GNNs~\cite{Specformer}.
We initialize a matrix $\mathbf{U}_{K}^{\prime} = [\mathbf{u}_1^{\prime}, \cdots, \mathbf{u}_{N^{\prime}}^{\prime}] \in \mathbb{R}^{N^{\prime} \times K}$ to match the principal eigenbasis of the real graph, denoted as $\mathbf{U}_{K} = [\mathbf{u}_1, \cdots, \mathbf{u}_{K_1}, \mathbf{u}_{N-K_2}, \cdots, \mathbf{u}_{N}] \in \mathbb{R}^{N \times K}$.

To eliminate the influence of GNNs, GDEM does not use the spectrum information during distillation.
Therefore, the first term in Equation~\ref{eq: L_GM} becomes:
\vspace{-5pt}
\begin{equation}
\label{eq: le}
    \mathcal{L}_{e} = \sum_{k=1}^{K} \left\| \mathbf{X}^{\top}\mathbf{u}_{k}\mathbf{u}_{k}^{\top} \mathbf{X} - {\mathbf{X}^{\prime}}^{\top}\mathbf{u}_{k}^{\prime}{\mathbf{u}^{\prime}_{k}}^{\top} \mathbf{X}^{\prime} \right\|_{F}^{2},
\end{equation}
where $\mathbf{u}_{k}\mathbf{u}_{k}^{\top}$ and $\mathbf{u}_{k}^{\prime}{\mathbf{u}^{\prime}_{k}}^{\top}$ are the subspaces induced by the $k$-th eigenvector in the real and synthetic graphs.

Additionally, as the basis of graph Fourier transform, eigenvectors are naturally normalized and orthogonal to each other.
However, directly optimizing $\mathbf{U}_{K}^{\prime}$ via gradient descent cannot preserve this property.
Therefore, an additional regularization is used to constrain the representation space:
\vspace{-5pt}
\begin{equation}
\label{eq: lo}
    \mathcal{L}_{o} = \left\| {\mathbf{U}_{K}^{\prime}}^{\top} \mathbf{U}_{K}^{\prime} - \mathbf{I}_{K} \right\|^2_{F}.
\end{equation}
See Appendix~\ref{app: implementation} for more implementation details.

\subsection{Discrimination Constraint}
\label{sec: discrimination}

In practice, we find that eigenbasis matching improves the cross-architecture generalization of GDEM but contributes less to the performance of node classification as it only preserves the global distribution, \ie, $\mathbf{X}^{\top}\mathbf{u}\mathbf{u}^{\top}\mathbf{X}$, without considering the information of downstream tasks.
Therefore, we need to approximate the second term in Equation~\ref{eq: L_GM}.
Interestingly, we find that $\mathbf{X}^{\top}\mathbf{A}\mathbf{Y} \in \mathbb{R}^{d \times C}$ indicates the category-level representations, which assigns each category a $d$-dimensional representation.
However, the MSE loss only emphasizes the intra-class similarity between the real and synthetic graphs and ignores the inter-class dissimilarity.

Based on this discovery, we design a discrimination constraint to effectively preserve the category-level information, which can also be treated as a class-aware regularization technique~\cite{IDM, CAFE}.
Specifically, we first learn the category-level representations of the real and synthetic graphs:
\vspace{-5pt}
\begin{equation}
    \mathbf{H} = \mathbf{Y}^{\top}\mathbf{A}\mathbf{X}, \quad
    \mathbf{H}^{\prime} =  {\mathbf{Y}^{\prime}}^{\top} \sum_{k=1}^{K}(1 - \lambda_{k})\mathbf{u}_{k}^{\prime}{\mathbf{u}_{k}^{\prime}}^{\top} \mathbf{X}^{\prime},
\end{equation}
where $\lambda_{k}$ is the $k$-th eigenvalue of the real graph Laplacian.
We then constrain the cosine similarity between $\mathbf{H}$ and $\mathbf{H}^{\prime}$:
\begin{equation}
\label{eq: ld}
    \mathcal{L}_{d} = \sum_{i=1}^{C}\left(1 - \frac{\mathbf{H}_{i}^{\top} \cdot \mathbf{H}_{i}^{\prime}}{||\mathbf{H}_{i}|| \: ||\mathbf{H}_{i}^{\prime}||} \right) + 
    \sum_{\substack{i,j=1 \\ i \ne j}}^{C} \frac{\mathbf{H}_{i}^{\top} \cdot \mathbf{H}_{j}^{\prime}}{||\mathbf{H}_{i}|| \: ||\mathbf{H}_{j}^{\prime}||}.
\end{equation}
% % Specifically, we aim to use the class-level representations of the synthetic graph as a classifier to categorize the nodes in the real graph.
% % However, the node features in a single subspace are insufficient since they only capture a portion of the structural information.
% % Therefore, we need to merge the node features from various subspaces:
% \begin{equation}
%     \overline{\mathbf{h}^{\prime}}_{c} = \sum_{k=1}^{K}(1 - \lambda_{k}) \overline{\mathbf{h}^{\prime}}_{c, k}, \quad
%     \overline{\mathbf{H}^{\prime}} = \left[\overline{\mathbf{h}^{\prime}}_{1}, \overline{\mathbf{h}^{\prime}}_2, \cdots, \overline{\mathbf{h}^{\prime}}_C\right]^{\top},
% \end{equation}
% where $\lambda_{k}$ is the $k$-th eigenvalue of the real graph Laplacian, and $\overline{\mathbf{H}^{\prime}} \in \mathbb{R}^{d \times C}$ is the learned classifier.
% Finally, we use it to classify the training nodes in the real graph:
% \begin{equation}
%      \mathcal{L}_{d} = - \sum_{i=1}^{N} \sum_{j=1}^{C} \mathbf{Y}_{ij} \ln{ \left( \mathbf{A} \mathbf{X} \overline{\mathbf{H}^{\prime}} \right)_{ij} },
% \end{equation}
% where $\mathbf{A} \mathbf{X} \overline{\mathbf{H}^{\prime}} \in \mathbb{R}^{N \times C}$ are the logits for classification.

Note that the discrimination constraint introduces the spectrum information in the distillation process, which conflicts with the eigenbasis matching.
However, we find that adjusting the weights of eigenbasis matching and the discrimination constraint can balance the performance and generalization of GDEM.
Ablation studies can be seen in Section~\ref{sec: ablation}.

\subsection{Final Objective and Synthetic Graph Construction}

In summary, the overall loss function of GDEM is formulated as the weighted sum of three regularization terms:
\begin{equation}
\label{eq: l_total}
    \mathcal{L}_{total} = \alpha \mathcal{L}_{e} + \beta \mathcal{L}_{d} + \gamma \mathcal{L}_{o},
\end{equation}
where $\alpha$, $\beta$, and $\gamma$ are the hyperparameters.
The pseudo-code of GDEM is presented in Algorithm~\ref{algo: pot}.

\begin{algorithm}[t]
\caption{GDEM for Graph Distillation}

\begin{algorithmic}
\label{algo: pot}
   \STATE {\bfseries Input:} Real graph $\mathcal{G}=\left ( \mathbf{A},\mathbf{X},\mathbf{Y} \right )$ with eigenvalues $\{\lambda_i\}_{i=1}^{K}$ and eigenbasis $\mathbf{U}_{K}$
   \STATE {\bfseries Init:} Synthetic graph $\mathcal{G}^\prime$ with eigenbasis $\mathbf{U}_{K}^{\prime}$, node features $\mathbf{X}^\prime$, and labels $\mathbf{Y}^{\prime}$
   \FOR{$t=1$ {\bfseries to} $T$}
   \STATE Compute $\mathcal{L}_e$, $\mathcal{L}_o$, and $\mathcal{L}_d$ via Eqs.~\ref{eq: le}, \ref{eq: lo}, and \ref{eq: ld}
   \STATE Compute $\mathcal{L}_{total} = \alpha \mathcal{L}_{e} + \beta \mathcal{L}_{d} + \gamma \mathcal{L}_{o}$
   \IF{$t\%(\tau _1+\tau _2)<\tau_1$}
   \STATE Update $\mathbf{U}_{K}^{\prime} \leftarrow \mathbf{U}_{K}^{\prime}-\eta _1\nabla_{\mathbf{U}_{K}^{\prime}}\mathcal{L}_{total}$
   \ELSE \STATE Update $\mathbf{X}^{\prime} \leftarrow \mathbf{X}^{\prime}-\eta _2\nabla_{\mathbf{X}^{\prime}}\mathcal{L}_{total}$
   \ENDIF
   \ENDFOR
   \STATE Compute $\mathbf{A}^{\prime} = \sum_{k=1}^{K} (1-\lambda_k) \mathbf{u}_{k}^{\prime} {\mathbf{u}_{k}^{\prime}}^{\top}$
   \STATE {\bfseries Return:} $\mathbf{A}^{\prime}$, $\mathbf{X}^{\prime}$
\end{algorithmic}

\end{algorithm}

Upon minimizing the total loss function, the outputs of GDEM are the eigenbasis and node features of the synthetic graph.
However, the data remains incomplete due to the absence of the graph spectrum.
Essentially, the graph spectrum encodes the global shape of a graph~\cite{Spectre}.
Ideally, if the synthetic graph preserves the distribution of the real graph, they should have similar spectrums.
Therefore, we directly replicate the real spectrum for the synthetic graph to construct its Laplacian matrix or adjacency matrix:
\vspace{-5pt}
\begin{equation}
    \mathbf{L}^{\prime} = \sum_{k=1}^{K} \lambda_k \mathbf{u}_{k}^{\prime} {\mathbf{u}_{k}^{\prime}}^{\top}, \quad
    \mathbf{A}^{\prime} = \sum_{k=1}^{K} (1-\lambda_k) \mathbf{u}_{k}^{\prime} {\mathbf{u}_{k}^{\prime}}^{\top}.
\end{equation}

\subsection{Discussion}

\paragraph{Complexity.}

The complexity of decomposition is $\mathcal{O}(N^{3})$. However, given that we only utilize the $K$ smallest or largest eigenvalues, the complexity reduces to $\mathcal{O}(KN^{2})$. Additionally, ${\mathbf{u}_{k}^{\top} \mathbf{X}}$ in Equation~\ref{eq: le} and $\mathbf{H}$ in Equation~\ref{eq: ld} cost $\mathcal{O}(KNd)$ and $\mathcal{O}(Ed)$ in pre-processing. During distillation, the complexity of $\mathcal{L}_e$, $\mathcal{L}_d$ and $\mathcal{L}_o$ are $\mathcal{O}(KN^{\prime}d+Kd^{2})$, $\mathcal{O}(KN^{\prime}d^{\prime}+Cd^2)$, and $\mathcal{O}(KN^{\prime2})$, respectively.

% The complexity of decomposition is $\mathcal{O}(N^{3})$. However, given that we only utilize the $K$ smallest or largest eigenvalues, the complexity reduces to $\mathcal{O}(KN^{2})$. Additionally, ${\mathbf{u}_{k}^{\top} \mathbf{X}}$ in Equation~\ref{eq: le} and $H$ in Equation~\ref{eq: ld} cost $\mathcal{O}(KNd)$ and $\mathcal{O}(Ed)$ in pre-processing.
% During distillation, the complexity of $\mathcal{L}_e$, $\mathcal{L}_d$ and $\mathcal{L}_o$ are $\mathcal{O}(K(N+N^{\prime})d)$, $\mathcal{O}(Cd^{2})$, and $\mathcal{O}(K{N^{\prime}}^{2})$, respectively.

% Additionally, Table~\ref{tab: complexity} presents the complexity of different GD methods. Additionally, Table~\ref{tab: complexity} presents the complexity of different GD methods. Additionally, Table~\ref{tab: complexity} presents the complexity of different GD methods.

\vspace{-10pt}
\paragraph{Relation to Message Passing.}
Message-passing (MP) is the most popular paradigm for GNNs.
Although GDEM does not explicitly perform message-passing during distillation, eigenbasis matching already encodes the information of neighbors as most MP operators rely on the combination of the out product of eigenvectors, \eg, $\mathbf{L}=\sum_{i=1}^{N}\lambda_{i}\mathbf{u}_{i}\mathbf{u}_{i}^{\top}$.
Therefore, GDEM not only inherits the expressive power of MP but also addresses the weaknesses of the previous distillation methods.

\vspace{-10pt}
\paragraph{Limitations.}
Hereby we discuss the limitations of GDEM.
(1) The decomposition of the real graph introduces additional computational costs for distillation.
(2) In scenarios with extremely high compression rates, the synthetic graphs can only match a limited number of real eigenbasis, resulting in performance degradation.

\vspace{-3pt}
\section{Theoretical Analysis}
\vspace{-3pt}
In this section, we give a theoretical analysis of GDEM and prove that it preserves the restricted spectral similarity.

\begin{definition}
(Spectral Similarity~\cite{RS})
% Given two matrices $\mathbf{A, B} \in \mathbb{R}^{n \times n}$, $\mathbf{B}$ is the $\epsilon$-spectral approximation of $\mathbf{A}$, if there exists a positive constant $\epsilon$, such that for all signals $\mathbf{s} \in \mathbb{R}^{n \times 1}$,
% \begin{equation}
%     (1-\epsilon)\mathbf{s}^{\top}\mathbf{A}\mathbf{s} <
%     \mathbf{s}^{\top}\mathbf{B}\mathbf{s} <
%     (1+\epsilon)\mathbf{s}^{\top}\mathbf{A}\mathbf{s}.
% \end{equation}
Let $\mathbf{A, B} \in \mathbb{R}^{N \times N}$ be two square matrices. Matrix $\mathbf{B}$ is considered a spectral approximation of $\mathbf{A}$ if there exists a positive constant $\epsilon$, such that for any vector $\mathbf{x} \in \mathbb{R}^{N}$, the following inequality holds:
\vspace{-5pt}
\begin{equation}
(1-\epsilon)\mathbf{x}^{\top}\mathbf{A}\mathbf{x} < \mathbf{x}^{\top}\mathbf{B}\mathbf{x} < (1+\epsilon)\mathbf{x}^{\top}\mathbf{A}\mathbf{x}.
\nonumber
\end{equation}
\end{definition}
\vspace{-5pt}
However, it is impossible to satisfy this condition for all $\mathbf{x}\in\mathbb{R}^{N}$~\cite{coarsening1}.
Therefore, we only consider a restricted version of spectral similarity in the feature space.
\begin{definition}
(Restricted Spectral Similarity, RSS \protect\footnotemark)
The synthetic graph Laplacian $\mathbf{L}^{\prime}$ preserves RSS of the real graph Laplacian $\mathbf{L}$, if there exists an $\epsilon>0$ such that:
\begin{equation}
    (1-\epsilon)\mathbf{x}^{\top}\mathbf{L}\mathbf{x} < {\mathbf{x}^{\prime}}^{\top} \mathbf{L}^{\prime} \mathbf{x}^{\prime} < (1+\epsilon)\mathbf{x}^{\top}\mathbf{L}\mathbf{x} 
    \quad \forall \mathbf{x}, \mathbf{x}^{\prime} \in \mathbf{X}, \mathbf{X}^{\prime}.
    \nonumber
\end{equation}
\end{definition}

\footnotetext{RSS defined in this paper is different from \citet{coarsening1}, which limits the signal $\mathbf{x}$ in the eigenvector space.}

% If the above inequality holds, the synthetic graph Laplacian $\mathbf{L}^\prime$ is an $\epsilon$-spectral approximation of the real graph Laplacian $\mathbf{L}$ and preserves its crucial structural information.

\begin{proposition}
The synthetic graph distilled by GDEM is a restricted $\epsilon$-spectral approximation of the real graph.
\label{prop}
\end{proposition}
\vspace{-15pt}
\begin{proof}
We first characterize the spectral similarity of node features in the real and synthetic graphs, respectively.
Notably, here we use the principal $K$ eigenvalues and eigenvectors as a truncated representation of the real graph
\vspace{-5pt}
\begin{align}
    \mathbf{x}^{\top}\mathbf{L}\mathbf{x} &= \mathbf{x}^{\top}\sum_{k=1}^{N} \lambda_{k} \mathbf{u}_{k}\mathbf{u}_{k}^{\top} \mathbf{x}
    \approx \sum_{k=1}^{K} \lambda_{k} \mathbf{x}^{\top}\mathbf{u}_{k}\mathbf{u}_{k}^{\top}\mathbf{x}, \label{eq: ss1} \\
    {\mathbf{x}^{\prime}}^{\top} \mathbf{L}^{\prime} \mathbf{x}^{\prime} &= {\mathbf{x}^{\prime}}^{\top} (\sum_{k=1}^{N^\prime} \lambda_{k} \mathbf{u}_{k}^{\prime} {\mathbf{u}_{k}^{\prime}}^{\top} + \Tilde{\mathbf{U}} \bm{\Lambda} \Tilde{\mathbf{U}}^{\top}) \mathbf{x}^{\prime} \nonumber \\
    &\approx \sum_{k=1}^{K} \lambda_{k} {\mathbf{x}^{\prime}}^{\top} \mathbf{u}_{k}^{\prime} {\mathbf{u}_{k}^{\prime}}^{\top} \mathbf{x}^{\prime} + \Delta,
\label{eq: ss2}
\end{align}
where $\Delta = {\mathbf{x}^{\prime}}^{\top} \Tilde{\mathbf{U}} \bm{\Lambda} \Tilde{\mathbf{U}}^{\top} \mathbf{x}^{\prime}$ and $\Tilde{\mathbf{U}}$ represents the non-orthogonal terms of the eigenbasis $\mathbf{U}^{\prime}_{K}$, which means that $\mathbf{U}^{\prime}_{K} + \Tilde{\mathbf{U}}$ is strictly orthogonal.

% Next, we characterize the spectral similarity of node features in a certain subspace $\mathbf{u}\mathbf{u}^{\top}$:
% \begin{equation}
%     \mathbf{x}^{\top}\mathbf{u}\mathbf{u}^{\top}\mathbf{x} = (\mathbf{u}\mathbf{u}^{\top}\mathbf{x})^{\top} (\mathbf{u}\mathbf{u}^{\top}\mathbf{x}) = \| \mathbf{u}\mathbf{u}^{\top}\mathbf{x} \|_{2}^{2}.
% \label{eq: uu_ss}
% \end{equation}

Combining Equations~\ref{eq: ss1} and \ref{eq: ss2}, we have
\vspace{-5pt}
\begin{equation}
\begin{aligned}
    &\left| \mathbf{x}^{\top}\mathbf{L}\mathbf{x}-{\mathbf{x}^{\prime}}^{\top} \mathbf{L}^{\prime} \mathbf{x}^{\prime} \right| \\
    \approx
    &\left| \sum_{k=1}^{K} \lambda_{k} \mathbf{x}^{\top}\mathbf{u}_{k}\mathbf{u}_{k}^{\top}\mathbf{x} - \sum_{k=1}^{K} \lambda_{k} {\mathbf{x}^{\prime}}^{\top} \mathbf{u}_{k}^{\prime} {\mathbf{u}_{k}^{\prime}}^{\top} \mathbf{x}^{\prime} - \Delta \right| \\
    \leq 
    &\underbrace{
    \sum_{k=1}^{K} \lambda_{k} \left| \mathbf{x}^{\top}\mathbf{u}_{k}\mathbf{u}_{k}^{\top}\mathbf{x} - {\mathbf{x}^{\prime}}^{\top} \mathbf{u}_{k}^{\prime} {\mathbf{u}_{k}^{\prime}}^{\top} \mathbf{x}^{\prime} \right|
    }_{\mathcal{L}_{e}} + 
    \underbrace{
    %
    % for height alignment
    \vphantom{\sum_{k=1}^{K} \lambda_{k} \left| \mathbf{x}^{\top}\mathbf{u}_{k}\mathbf{u}_{k}^{\top}\mathbf{x} - {\mathbf{x}^{\prime}}^{\top} \mathbf{u}_{k}^{\prime} {\mathbf{u}_{k}^{\prime}}^{\top} \mathbf{x}^{\prime} \right|}
    \left| \Delta \right|
    }_{\mathcal{L}_{o}}.
\label{eq: bound}
\end{aligned}
\end{equation}
The above inequality shows that the objective of eigenbasis matching is the upper bound of the spectral discrepancy between the real and synthetic graphs.
Optimizing $\mathcal{L}_{e}$ and $\mathcal{L}_{o}$ makes the bound tighter and preserves the spectral similarity of the real graph. 
The synthetic graph is a restricted $\epsilon$-spectral approximation of the real graph with $\epsilon = \sum_{k=1}^{K} \lambda_{k}\left| \mathbf{x}^{\top}\mathbf{u}_{k}\mathbf{u}_{k}^{\top}\mathbf{x} - {\mathbf{x}^{\prime}}^{\top} \mathbf{u}_{k}^{\prime} {\mathbf{u}_{k}^{\prime}}^{\top} \mathbf{x}^{\prime} \right| +  \left| \Delta \right|$.
\end{proof}

\section{Experiments}

In this section, we conduct experiments on a variety of graph datasets to validate the effectiveness, generalization, and efficiency of the proposed GDEM.

\vspace{-5pt}
\subsection{Experimental Setup}

\begin{table*}[ht]
  \centering
  \caption{Node classification performance of different distillation methods, mean accuracy (\%) ± standard deviation. \textbf{Bold} indicates the best performance and \underline{underline} means the runner-up.}
  \resizebox{\linewidth}{!}{
    \begin{tabular}{lcccccccccc}
    \toprule
    \multirow{2}[4]{*}{\textbf{Dataset}} & \multirow{2}[4]{*}{\textbf{Ratio ($r$)}} & \multicolumn{4}{c}{\textbf{Traditional Methods}} & \multicolumn{4}{c}{\textbf{Graph Distillation Methods}} & \multirow{2}[4]{*}{\makecell{\textbf{Whole} \\ \textbf{Dataset}}} \\
    
    \cmidrule(lr){3-6} \cmidrule(lr){7-10} &       & \makecell{Random \\ $(\mathbf{A}^{\prime}, \mathbf{X}^{\prime})$} & \makecell{Coarsening \\ $(\mathbf{A}^{\prime}, \mathbf{X}^{\prime})$} & \makecell{Herding \\ $(\mathbf{A}^{\prime}, \mathbf{X}^{\prime})$} & \makecell{K-Center \\ $(\mathbf{A}^{\prime}, \mathbf{X}^{\prime})$} & \makecell{GCOND \\ $(\mathbf{A}^{\prime}, \mathbf{X}^{\prime})$} & \makecell{SFGC \\ $(\mathbf{X}^{\prime})$}  & \makecell{SGDD \\ $(\mathbf{A}^{\prime}, \mathbf{X}^{\prime})$} & \makecell{GDEM \\ $(\mathbf{U}^{\prime}, \mathbf{X}^{\prime})$} &  \\
    \midrule
    \multirow{3}[2]{*}{Citeseer} & 0.90\% & 54.4±4.4 & 52.2±0.4 & 57.1±1.5 & 52.4±2.8 & 70.5±1.2 & \underline{71.4±0.5} & 69.5±0.4 & \textbf{72.3±0.3} & \multirow{3}[2]{*}{71.7±0.1} \\
          & 1.80\% & 64.2±1.7 & 59.0±0.5 & 66.7±1.0 & 64.3±1.0 & 70.6±0.9 & \underline{72.4±0.4} & 70.2±0.8 & \textbf{72.6±0.6} &  \\
          & 3.60\% & 69.1±0.1 & 65.3±0.5 & 69.0±0.1 & 69.1±0.1 & 69.8±1.4 & \underline{70.6±0.7} & 70.3±1.7 & \textbf{72.6±0.5} &  \\
    \midrule
    \multirow{3}[2]{*}{Pubmed} & 0.08\% & 69.4±0.2 & 18.1±0.1 & \underline{76.7±0.7} & 64.5±2.7 & 76.5±0.2 & 76.4±1.2 & 77.1±0.5 & \textbf{77.7±0.7} & \multirow{3}[2]{*}{79.3±0.2} \\
          & 0.15\% & 73.3±0.7 & 28.7±4.1 & 76.2±0.5 & 69.4±0.7 & 77.1±0.5 & \underline{77.5±0.4} & 78.0±0.3 & \textbf{78.4±1.8} &  \\
          & 0.30\% & 77.8±0.3 & 42.8±4.1 & 78.0±0.5 & \textbf{78.2±0.4} & 77.9±0.4 & 77.9±0.3 & 77.5±0.5 & \textbf{78.2±0.8} &  \\
    \midrule
    \multirow{3}[2]{*}{Ogbn-arxiv} & 0.05\% & 47.1±3.9 & 35.4±0.3 & 52.4±1.8 & 47.2±3.0 & 59.2±1.1 & \textbf{65.5±0.7} & 60.8±1.3 & \underline{63.7±0.8} & \multirow{3}[2]{*}{71.4±0.1} \\
          & 0.25\% & 57.3±1.1 & 43.5±0.2 & 58.6±1.2 & 56.8±0.8 & 63.2±0.3 & \textbf{66.1±0.4} & \underline{65.8±1.2} & 63.8±0.6 &  \\
          & 0.50\% & 60.0±0.9 & 50.4±0.1 & 60.4±0.8 & 60.3±0.4 & 64.0±0.4 & \textbf{66.8±0.4} & \underline{66.3±0.7} & 64.1±0.3 &  \\
    \midrule
    \multirow{3}[2]{*}{Flickr} & 0.10\% & 41.8±2.0 & 41.9±0.2 & 42.5±1.8 & 42.0±0.7 & 46.5±0.4 & 46.6±0.2 & \underline{46.9±0.1} & \textbf{49.9±0.8} & \multirow{3}[2]{*}{47.2±0.1} \\
          & 0.50\% & 44.0±0.4 & 44.5±0.1 & 43.9±0.9 & 43.2±0.1 & \underline{47.1±0.1} & 47.0±0.1 & \underline{47.1±0.3} & \textbf{49.4±1.3} &  \\
          & 1.00\% & 44.6±0.2 & 44.6±0.1 & 44.4±0.6 & 44.1±0.4 & \underline{47.1±0.1} & \underline{47.1±0.1} & \underline{47.1±0.1} & \textbf{49.9±0.6} &  \\
    \midrule
    \multirow{3}[2]{*}{Reddit} & 0.05\% & 46.1±4.4 & 40.9±0.5 & 53.1±2.5 & 46.6±2.3 & 88.0±1.8 & 89.7±0.2 & \underline{91.8±1.9} & \textbf{92.9±0.3} & \multirow{3}[2]{*}{93.9±0.0} \\
          & 0.10\% & 58.0±2.2 & 42.8±0.8 & 62.7±1.0 & 53.0±3.3 & 89.6±0.7 & 90.0±0.3 & \underline{91.0±1.6} & \textbf{93.1±0.2} &  \\
          & 0.50\% & 66.3±1.9 & 47.4±0.9 & 71.0±1.6 & 58.5±2.1 & 90.1±0.5 & 89.9±0.4 & \underline{91.6±1.8} & \textbf{93.2±0.4} &  \\
    \midrule
    \multirow{3}[2]{*}{Squirrel} & 0.60\% & 22.4±1.6 & 20.9±1.1 & 21.3±1.1 & 21.8±0.3 & \underline{27.0±1.3} & 24.0±0.4 & 24.1±2.3 & \textbf{28.4±2.0} & \multirow{3}[2]{*}{33.0±0.4} \\
          & 1.20\% & 25.0±0.2 & 21.1±0.4 & 21.4±2.1 & 22.8±0.9 & 25.7±2.3 & \underline{26.9±2.5} & 24.7±2.5 & \textbf{28.2±2.4} &  \\
          & 2.50\% & \underline{26.9±1.4} & 21.5±0.3 & 22.4±1.6 & 22.9±1.7 & 25.3±0.8 & 26.1±0.8 & 25.8±1.8 & \textbf{27.8±1.6} &  \\
    \midrule
    \multirow{3}[2]{*}{Gamers} & 0.05\% & 56.6±1.8 & 56.1±0.1 & 56.7±1.7 & 52.5±4.2 & \underline{58.5±1.5} & 58.2±1.1 & 57.5±1.8 & \textbf{59.3±1.9} & \multirow{3}[2]{*}{62.6±0.0} \\
          & 0.25\% & \underline{60.5±1.0} & 56.9±3.0 & 57.5±2.0 & 57.2±2.3 & 58.9±1.8 & 58.8±0.5 & 57.7±1.0 & \textbf{60.8±0.4} &  \\ 
          & 0.50\% & \underline{60.0±0.5} & 57.1±0.4 & 58.6±1.3 & 57.8±1.7 & 58.5±1.9 & 59.9±0.3 & 58.4±1.7 & \textbf{61.2±0.3} &  \\
    \bottomrule
    \end{tabular}}
  \label{tab: effectiveness}
\end{table*}

\noindent \textbf{Datasets.}
To evaluate the effectiveness of our GDEM, we select seven representative graph datasets, including five homophilic graphs, \ie, Citeseer, Pubmed~\cite{GCN}, Ogbn-arxiv~\cite{OGB}, Filckr~\cite{Flickr}, and Reddit~\cite{SAGE}, and two heterophilic graphs, \ie, Squirrel~\cite{squirrel} and Gamers~\cite{gamers}.

\noindent \textbf{Baselines.}
We benchmark our model against several competitive baselines, which can be divided into two categories:
(1) Traditional graph reduction methods, including three coreset methods, \ie, Random, Herding, and K-Center~\cite{core_set1, core_set2}, and one coarsening method~\cite{coarsening1}.
(2) Graph distillation methods, including two gradient matching methods, \ie, GCOND~\cite{GCond} and SGDD~\cite{SGDD}, and one trajectory matching method, \ie, SFGC~\cite{SFGC}.
See Appendix~\ref{app: baseline} for more details.

\noindent \textbf{Evaluation Protocol.}
To fairly evaluate the quality of synthetic graphs, we perform the following two steps for all methods:
(1) \textbf{Distillation step}, where we apply the distillation methods in the training set of the real graphs.
(2) \textbf{Evaluation step}, where we train GNNs on the synthetic graph from scratch and then evaluate their performance on the test set of real graphs.
In the node classification experiment (Section~\ref{sec: classification}), we follow the settings of the original papers~\cite{GCond, SFGC, SGDD}.
In the generalization experiment (Section~\ref{sec: generalization}), we use six representative GNNs, including three spatial GNNs, \ie, GCN, SGC, and PPNP, and three spectral GNNs, \ie, ChebyNet, BernNet, and GPR-GNN.
See Appendix~\ref{app: evaluation} for more detailed description.

\noindent \textbf{Settings and Hyperparameters.}
To eliminate randomness, in the distillation step, we run the distillation methods 10 times and yield 10 synthetic graphs. Moreover, we set $K_1+K_2=N^{\prime}$. To reduce the tuning complexity, we treat $r_k=\left\{0.8, 0.85, 0.9, 0.95, 1.0\right\}$ as a hyperparameter and set $K_1=r_kN^\prime$, $K_2=(1-r_k)N^\prime$ for eigenbasis matching.
In the evaluation step, spatial GNNs have two aggregation layers and the polynomial order of spectral GNNs is set to 10. For more details, see Appendix~\ref{app: hyperparameter}.

\subsection{Node Classification}
\label{sec: classification}

The node classification performance is reported in Table~\ref{tab: effectiveness}, in which we have the following observations:

First, the GD methods consistently outperform the traditional methods, including coreset and coarsening.
The reasons are two-fold: On the one hand, GD methods can leverage the powerful representation learning ability of GNNs to synthesize the graph data. On the other hand, the distillation process involves the downstream task information. In contrast, the traditional methods can only leverage the structural information.

\begin{table*}[t]
  \centering
  \caption{Generalization of different distillation methods across GNNs. $\uparrow$ means higher the better and $\downarrow$ means lower the better. \textmd{Avg.}, \textmd{Std.}, and \textmd{Impro.} indicate average accuracy, standard deviation, and absolute performance improvement.}
  \resizebox{0.9\linewidth}{!}{
    \begin{tabular}{clccccccccc}
    \toprule
    \multirow{2}[4]{*}{\makecell{\textbf{Dataset}\\(Ratio)}} & \multirow{2}[4]{*}{\textbf{Methods}} & \multicolumn{3}{c}{\textbf{Spatial GNNs}} & \multicolumn{3}{c}{\textbf{Spectral GNNs}} & \multirow{2}[4]{*}{\textbf{Avg.} ($\uparrow$)} & \multirow{2}[4]{*}{\textbf{Std.} ($\downarrow$)} & \multirow{2}[4]{*}{\textbf{Impro.} ($\uparrow$)} \\
    \cmidrule(lr){3-5} \cmidrule(lr){6-8}         &       & GCN   & SGC   & PPNP & ChebyNet & BernNet & GPR-GNN &       &  \\
    \midrule
    \multirow{3}[2]{*}{\makecell{Citeseer\\($r=1.80\%$)}} & GCOND & 70.5 & 70.3 & 69.6 & 68.3 & 63.1 & 67.2 & 68.17 & 2.54 & (+) 4.21\\
    & SFGC  & 71.6 & 71.8 & 70.5 & 71.8 & 71.1 & 71.7 & 71.42 & \textbf{0.47} & (+) 0.96 \\
    & SGDD  & 70.2 & 71.3 & 69.2 & 70.5 & 64.7 & 69.7 & 69.27 & 2.14 & (+) 3.11 \\
    & GDEM  & 72.6 & 72.1 & 72.6 & 71.4 & 72.6 & 73.0 & \textbf{72.38} & 0.51 & - \\
    \midrule
    \multirow{3}[2]{*}{\makecell{Pubmed\\($r=0.15\%$)}} & GCOND & 77.7 & 77.6 & 77.3 & 76.0 & 74.4 & 76.5 & 76.58 & 1.15 & (+) 1.34 \\
    & SFGC  & 77.5 & 77.4 & 77.6 & 77.3 & 76.4 & 78.6 & 77.47 & \textbf{0.64} & (+) 0.45 \\
    & SGDD  & 78.0 & 76.6 & 78.7 & 76.9 & 75.5 & 77.0 & 77.12 & 1.02 & (+) 0.80 \\
    & GDEM  & 78.4 & 76.1 & 78.1 & 78.1 & 78.2 & 78.6 & \textbf{77.92} & 0.83 & - \\
    \midrule
    \multirow{3}[2]{*}{\makecell{Ogbn-arxiv\\($r=0.25\%$)}} & GCOND & 63.2 & 63.7 & 63.4 & 54.9 & 55.0 & 60.5 & 60.12 & 3.80 & (+) 2.90 \\
    & SFGC  & 65.1 & 64.8 & 63.9 & 60.7 & 63.8 & 64.9 & \textbf{63.87} & 1.50 & (-) 0.85 \\
    & SGDD  & 65.8 & 64.0 & 63.6 & 56.4 & 62.0 & 64.0 & 62.63 & 3.00 & (+) 0.39 \\
    & GDEM  & 63.8 & 62.9 & 63.5 & 62.4 & 61.9 & 63.6 & 63.02 & \textbf{0.69} & - \\
    \midrule
    \multirow{3}[2]{*}{\makecell{Flickr\\($r=0.50\%$)}} & GCOND & 47.1 & 46.1 & 45.9 & 42.8 & 44.3 & 46.4 & 45.43 & 1.45 & (+) 3.90 \\
    & SFGC  & 47.1 & 42.5 & 40.7 & 45.4 & 45.7 & 46.4 & 44.63 & 2.27 & (+) 4.70 \\
    & SGDD  & 47.1 & 46.5 & 44.3 & 45.3 & 46.0 & 46.8 & 46.00 & 0.96 & (+) 3.33 \\
    & GDEM  & 49.4 & 50.3 & 49.4 & 48.3 & 49.6 & 49.0 & \textbf{49.33} & \textbf{0.60} & - \\
    \midrule
    \multirow{3}[2]{*}{\makecell{Reddit\\($r=0.10\%$)}} & GCOND & 89.4 & 89.6 & 87.8 & 75.5 & 67.1 & 78.8 & 81.37 & 8.35 & (+) 10.10 \\
    & SFGC  & 89.7 & 89.5 & 88.3 & 82.8 & 87.8 & 85.4 & 87.25 & 2.44 & (+) 4.22 \\
    & SGDD  & 91.0 & 89.4 & 89.2 & 78.4 & 72.4 & 81.4 & 83.63 & 6.80 & (+) 7.84 \\
    & GDEM  & 93.1 & 90.0 & 92.6 & 90.0 & 92.7 & 90.4 & \textbf{91.47} & \textbf{1.35} & - \\
    \midrule
    \multirow{3}[2]{*}{\makecell{Squirrel\\($r=1.20\%$)}} & GCOND & 25.7 & 27.2 & 23.2 & 23.3 & 26.0 & 26.6 & 25.33 & 1.55 & (+) 1.89 \\
    & SFGC  & 26.9 & 24.2 & 27.2 & 25.3 & 25.5 & 26.6 & 25.95 & \textbf{1.04} & (+) 1.27 \\
    & SGDD  & 24.7 & 27.2 & 22.4 & 24.5 & 24.7 & 27.3 & 25.13 & 1.69 & (+) 2.09 \\
    & GDEM  & 28.2 & 28.0 & 25.4 & 26.1 & 28.2 & 27.4 & \textbf{27.22} & 1.09 & - \\
    \midrule
    \multirow{3}[2]{*}{\makecell{Gamers\\($r=0.25\%$)}} & GCOND & 58.9 & 54.2 & 60.1 & 60.3 & 59.1 & 59.3 & 58.65 & 2.05 & (+) 1.57 \\
    & SFGC  & 58.8 & 55.0 & 56.3 & 57.2 & 57.5 & 59.8 & 57.43 & 1.57 & (+) 2.79 \\
    & SGDD  & 57.7 & 54.6 & 56.0 & 57.3 & 58.8 & 58.6 & 57.17 & 1.47 & (+) 3.05 \\
    & GDEM  & 60.8 & 59.5 & 61.0 & 59.9 & 59.8 & 60.3 & \textbf{60.22} & \textbf{0.54} & - \\
    \bottomrule
    \end{tabular}}
\label{tab: generalization}
\end{table*}

Second, GDEM achieves state-of-the-art performance in 6 out of 7 graph datasets, demonstrating its effectiveness in preserving the distribution of real graphs.
Existing GD methods heavily rely on the information of GNNs to distill synthetic graphs.
However, the results of GDEM reveal that matching eigenbasis can also yield good synthetic graphs.
Furthermore, some results of GDEM are better than those on the entire dataset, which may be due to the use of high-frequency information.

Third, GDEM performs slightly worse on Ogbn-arxiv but achieves promising results on other large-scale graphs. We conjecture this is because, under the compression ratios of 0.05\% - 0.50\%, there are only hundreds of eigenvectors for eigenbasis matching, which is not enough to cover all the useful subspaces in Ogbn-arxiv. See Appendix~\ref{app: analysis of arXiv} for further experimental verification. 

% The reason is that arXiv is a large-scale graph data, which has nearly 0.17M nodes. Under the compression ratios of 0.05\% - 0.50\%, there are only hundreds of eigenvectors for eigenbasis matching, which cannot cover all the useful subspaces.

\subsection{Cross-architecture Generalization}
\label{sec: generalization}

We evaluate the generalization ability of the synthetic graphs distilled by four different GD methods, including GCOND, SFGC, SGDD, and GDEM.
In particular, each synthetic graph is evaluated by six GNNs, and the average accuracy and variance of the evaluation results are shown in Table~\ref{tab: generalization}.
% We have the following discoveries:

First, GDEM stands out by exhibiting the highest average accuracy across datasets except for Ogbn-arxiv, indicating that the synthetic graphs distilled by GDEM can consistently benefit a variety of GNNs.
Moreover, GDEM significantly reduces the performance gap between different GNNs. For example, the variance of GCOND is 2-6 times higher than that of GDEM.
On the other hand, SGDD broadcasts the structural information to synthetic graphs and exhibits better generalization ability than GCOND, implying that preserving graph structures can improve the generalization of synthetic graphs.
SFGC proposes structure-free distillation.
However, this strategy may lead to restricted application scenarios due to the lack of explicit graph structures.

% Second, comparing the generalization results of the two gradient matching methods, \ie, GCOND and SFGC, we can find that compressing the graph structure into node features can narrow the performance gap between different GNNs.
% However, this approach is less suitable for spectral GNNs, because it treats the identity matrix as the adjacency matrix.
% Generally, spectral GNNs are designed to learn the coefficients associated with different orders of the adjacency matrix~\cite{GPR-GNN, BernNet}.
% Unfortunately, training on the identity matrix does not facilitate the acquisition of valuable coefficients.

\subsection{Optimal Performance and Time Overhead}
\label{sec: optimal}

% \begin{table*}[t]
%   \centering
%   \caption{The graph classification performance comparison to baselines, mean accuracy (\%) or ROC-AUC ± standard deviation. \textbf{Bold} indicates the best performance and \underline{underline} means the runner-up.}
%     \begin{tabular}{ccccccc}
%     \toprule
%     \multirow{2}[0]{*}{\textbf{Dataset}} & \multirow{2}[0]{*}{\textbf{Graphs/Cls.}} & \multirow{2}[0]{*}{\textbf{Ratio}} & \multicolumn{3}{c}{\textbf{Methods}} & \multirow{2}[0]{*}{\textbf{Whole Dataset}} \\
%     \cmidrule(lr){4-6}  &   &   & DosCond & KiDD & GDEM & \\
%       \midrule
%     \multirow{1}[2]{*}{\makecell{DD\\(Accuracy)}} & 1 & 0.20\% & \underline{70.4±2.2} & 69.7±1.0 & \textbf{72.6±1.1} & \multirow{2}[0]{*}{78.9±0.6} \\
%       & 10 & 2.10\% & \underline{73.5±1.1} & 71.1±0.8 & \textbf{74.0±1.1} &  \\
%       \midrule
%     \multirow{1}[2]{*}{\makecell{ogbg-molbbbp\\(ROC-AUC)}} & 1 & 0.10\% & 0.581±0.005 & \textbf{0.628±0.011} & \underline{0.618±0.019} & \multirow{2}[0]{*}{0.646±0.004} \\
%       & 10 & 1.20\% & 0.605±0.008 & \underline{0.644±0.011} & \textbf{0.668±0.024} &  \\
%       \bottomrule
%     \end{tabular}%
%   \label{tab: graph classification}%
% \end{table*}%

We compare the optimal performance and time overhead of different GD methods by traversing various GNN architectures in the Pubmed dataset.
% The results are shown in Tables~\ref{tab: optimal} and \ref{tab: overhead}.
Since GDEM does not use GNNs during distillation, we remove the inner- and outer-loop of GCOND and SGDD when calculating the time overhead for a fair comparison.
Therefore, the running time is faster than the results in~\citet{SGDD}.

\begin{table}[t]
  \centering
  \caption{Optimal performance of different methods.}
    \resizebox{\linewidth}{!}{
    \begin{tabular}{lcccccc}
        \toprule
        Evaluation & GCN & SGC & \multicolumn{1}{c}{PPNP} & Cheb. & Bern. & GPR. \\
        \midrule
        GCOND   & 77.7 & \textbf{77.6} & 77.9 & 77.3 & \textbf{78.2} & 78.3 \\
        % SFGC    &  &  &  &  &  &  \\ 
        SGDD    & 78.0 & 76.6 & \textbf{78.7} & 77.5 & 78.0 & 78.3 \\ 
        GDEM    & \textbf{78.4} & 76.1 & 78.1 & \textbf{78.1} & \textbf{78.2} & \textbf{78.6} \\
        \bottomrule
    \end{tabular}}
    \label{tab: optimal}
\end{table}

\begin{table}[t]
    \centering
    \caption{Time overhead ($s$) of different methods.}
    \resizebox{\linewidth}{!}{
    \begin{tabular}{lccccccc}
    \toprule
    Distillation & GCN & SGC & PPNP & Cheb. & Bern. & GPR. & Overall \\
    % \midrule
    % GCOND & 1.99±0.48 & 1.36±0.48 & 1.52±0.49 & 3.89±0.40 & 56.94±0.54 & 3.05±0.49 \\
    % SGDD  & 2.95±0.44 & 2.18±0.33 & 2.33±0.57 & 4.95±0.49 & 58.07±0.41 & 4.28±0.43 \\
    \midrule
    GCOND & 1.99 & 1.36 & 1.52 & 3.89 & 56.94 & 3.05 & 68.75\\
    SGDD  & 2.95 & 2.18 & 2.33 & 4.95 & 58.07 & 4.28 & 74.76 \\
    \midrule
    GDEM  & - & - & - & - & - & - & \textbf{1.79} \\
    \bottomrule
    \end{tabular}}
    \label{tab: overhead}
\end{table}

In Table~\ref{tab: optimal}, we can find that both GCOND and SGDD improve their performance by traversing different GNNs compared to the results in Table~\ref{tab: generalization}. However, this strategy also introduces additional computation costs in the distillation stage.
As shown in Table~\ref{tab: overhead}, the complexity of GCOND and SGDD is related to the complexity of distillation GNNs.
Notably, when choosing GNNs with high complexity, \eg, BernNet, their time overhead will increase significantly.
On the other hand, GDEM still exhibits remarkable performance compared to the traversal results of GCOND and SGDD.
More importantly, the complexity of GDEM will not be affected by GNNs, which eliminates the traversal requirement of previous methods.
As a result, the overall time overhead of GDEM is significantly smaller than GCOND and SGDD, which validates the efficiency of GDEM. See Appendix~\ref{app: cross} for more generalization results of GCOND and SGDD.

% \subsection{Graph Classification}
% \textcolor{blue}{We also investigate whether GDEM works well in graph classification. Specifically, we conduct experiments on DD~\cite{morris2020tudataset} and ogbg-molbbbp~\cite{hu2020open},, and choose DosCond~\cite{jin2022condensing} and KiDD~\cite{xu2023kernel} as our baselines. For DD, we randomly split the graphs into 80\%, 10\%, and 10\% for training, validation, and testing. For ogbg-molbbbp, we apply the public splits of OGB~\cite{hu2020open}. We use GCN as the backbone GNN for graph classification. The graph classification results are listed in Table~\ref{tab: graph classification}. }

% \textcolor{blue}{From the table, we observe that GDEM achieves state-of-the-art performance in 3 out of 4 cases, demonstrating its effectiveness in graph-level tasks. It can be seen that the performance of GDEM is significantly comparable to the original datasets. Specifically, GDEM approximates the performance of original training by 93.8\% with 2.10\% graphs on DD, and even achieves the lossless distillation on ogbg-molbbbp.}

\subsection{Ablation Study}
\label{sec: ablation}

We perform ablation studies in the Pubmed and Gamers datasets to verify the effectiveness of different regularization terms, \ie, $\mathcal{L}_{e}$, $\mathcal{L}_{o}$, and $\mathcal{L}_{d}$.

\vspace{-8pt}
\paragraph{Model Analysis.}
Table~\ref{tab: ab} shows the roles of different regularization terms.
First, all of them contribute to both the effectiveness and generalization of GDEM.
Specifically, $\mathcal{L}_{e}$ and $\mathcal{L}_{o}$ primarily govern the generalization ability of GDEM, as the variance of GNNs increases significantly when removing either of them.
Second, we observe that $\mathcal{L}_{d}$ hurts the generalization of GDEM. The reason is that the discrimination constraint uses the information of the graph spectrum and introduces the low-frequency preference. But it also improves the performance of GDEM.
Therefore, GDEM needs to carefully balance these two loss functions.
\begin{table}[t]
  \centering
  \caption{Ablation studies on Pubmed / Gamers.}
    \resizebox{\linewidth}{!}{
    \begin{tabular}{lcccc}
    \toprule
    Pubmed
    & GCN ($\uparrow$)  & GPR. ($\uparrow$) & Avg. ($\uparrow$) & Var. ($\downarrow$) \\
    \midrule
    GDEM  & 78.4 / 60.8 & 78.6 / 60.3 & 77.92 / 60.22 & 0.69 / 0.29 \\
    w/o $\mathcal{L}_{e}$ & 76.1 / 56.5 & 76.9 / 59.8 & 76.13 / 58.93 & 1.18 / 2.39 \\
    w/o $\mathcal{L}_{o}$ & 77.9 / 59.0 & 76.4 / 58.9 & 77.07 / 58.85 & 2.15 / 2.34 \\
    w/o $\mathcal{L}_{d}$ & 76.7 / 59.9 & 77.2 / 60.3 & 76.77 / 59.78 & 0.21 / 0.13 \\
    \bottomrule
    \end{tabular}}
  \label{tab: ab}
\end{table}

\begin{figure}[t]
    \centering
    \subfigure[Analysis of $\alpha$ with $\beta$=\num{e-5}]{
        \centering
        \includegraphics[width=0.45\linewidth]{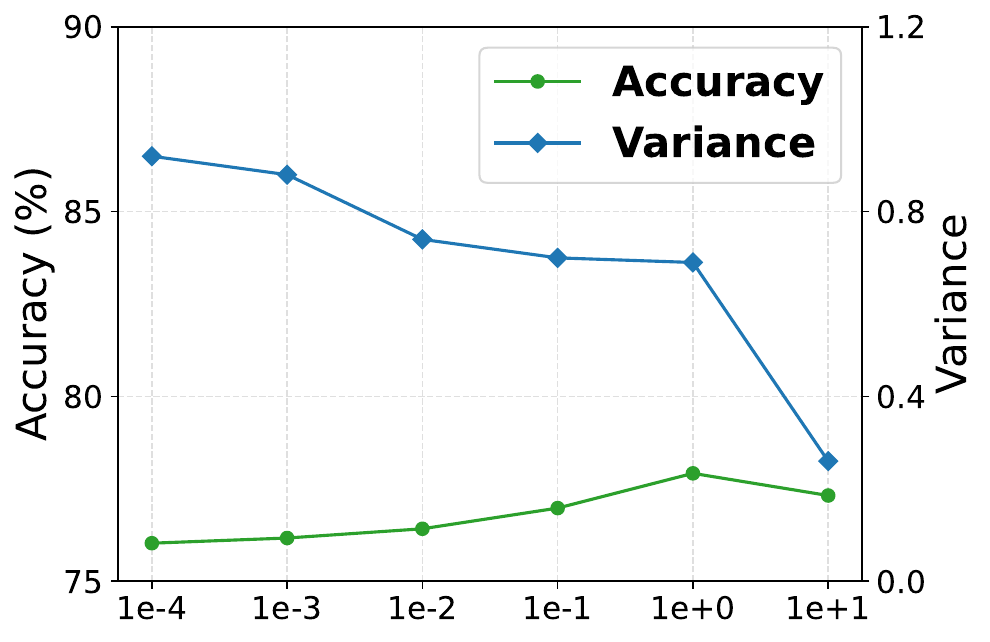}
        \label{fig: alpha}
    }
    \subfigure[Analysis of $\beta$ with $\alpha$=$1.0$]{
        \includegraphics[width=0.45\linewidth]{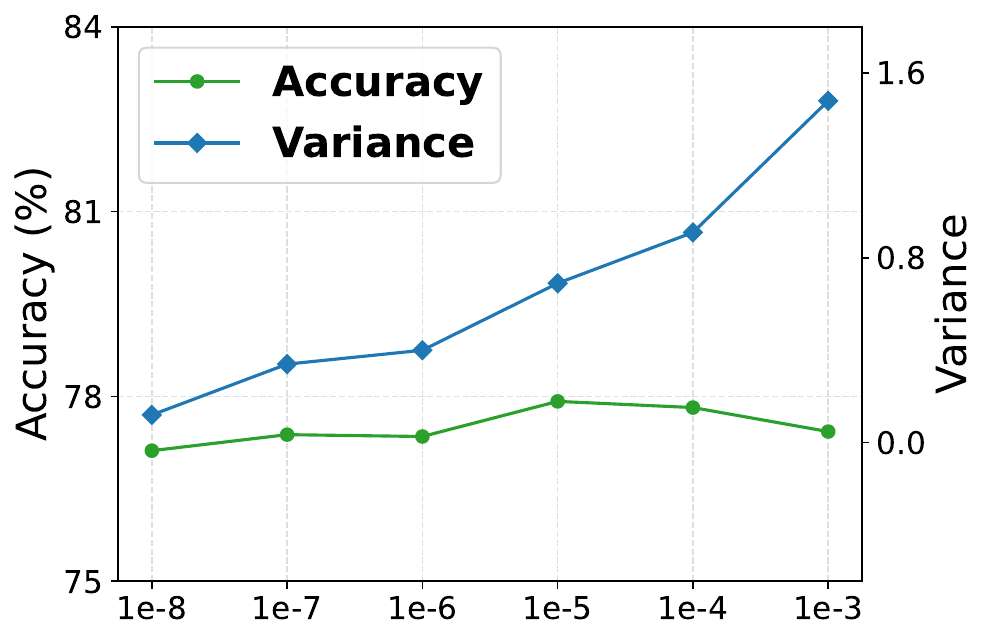}
        \label{fig: beta}
    }
    \caption{Influence of $\mathcal{L}_e$ and $\mathcal{L}_d$ in GDEM.}
    \label{fig: parameter}
\end{figure}

\vspace{-8pt}
\paragraph{Parameters Analysis.}
We conduct an additional parameter analysis to further demonstrate the influence of $\mathcal{L}_e$ and $\mathcal{L}_d$, as illustrated in Figure~\ref{fig: parameter}.
Specifically, we observe that with the increase in $\alpha$, the variance of GDEM gradually decreases.
However, a higher value of $\alpha$ also leads to performance degeneration.
On the other hand, increasing the value of $\beta$ will continue to increase the variance of GDEM but the accuracy decreases when $\beta$ surpasses a specific threshold.

\subsection{Visualization}

\begin{figure}[t]
    \centering
    \subfigure[GCOND]{
        \centering
        \includegraphics[width=0.3\linewidth]{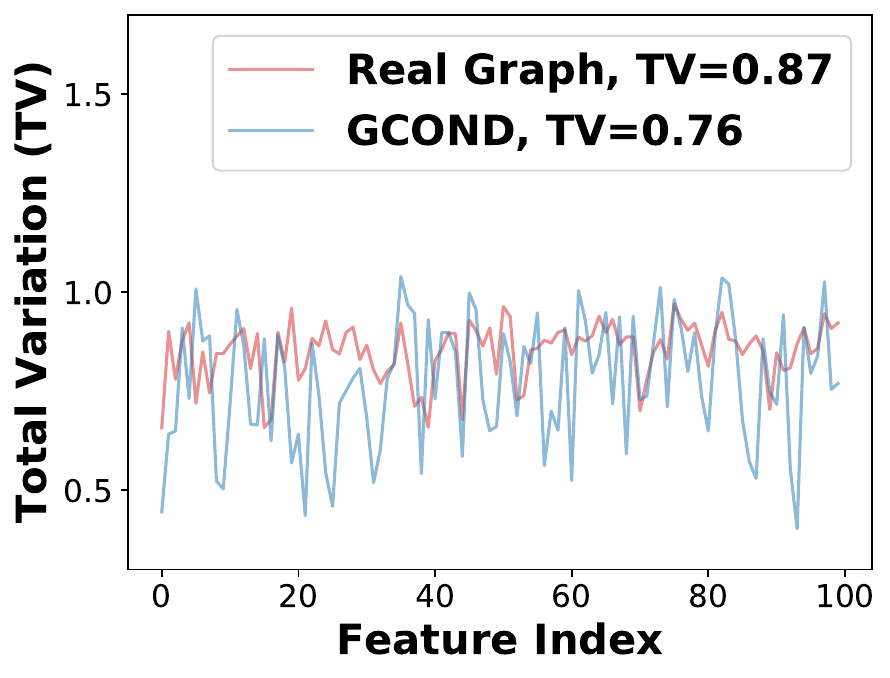}
    }
    \subfigure[SGDD]{
        \centering
        \includegraphics[width=0.3\linewidth]{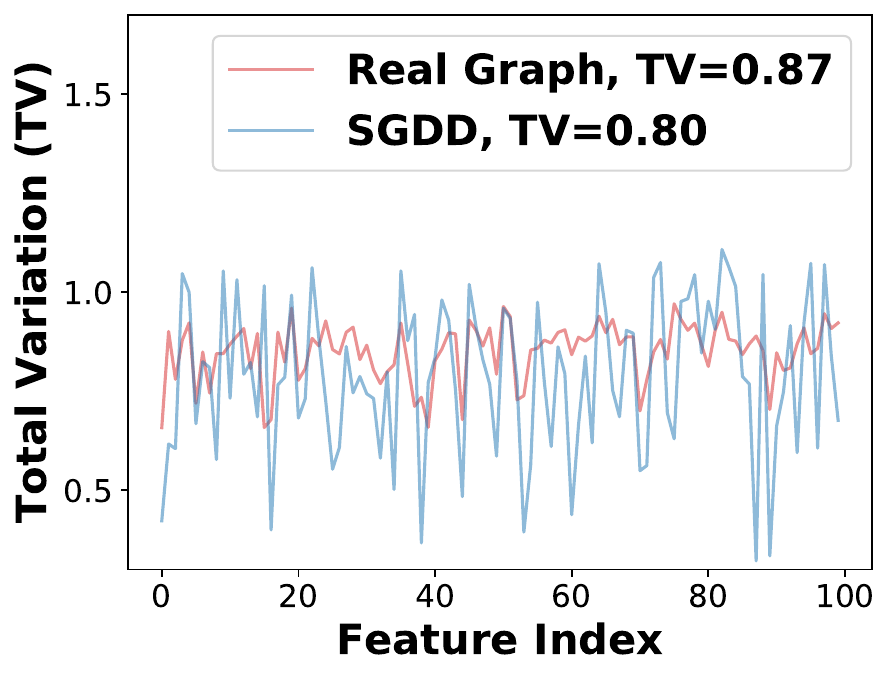}
    }
    \subfigure[GDEM]{
        \centering
        \includegraphics[width=0.3\linewidth]{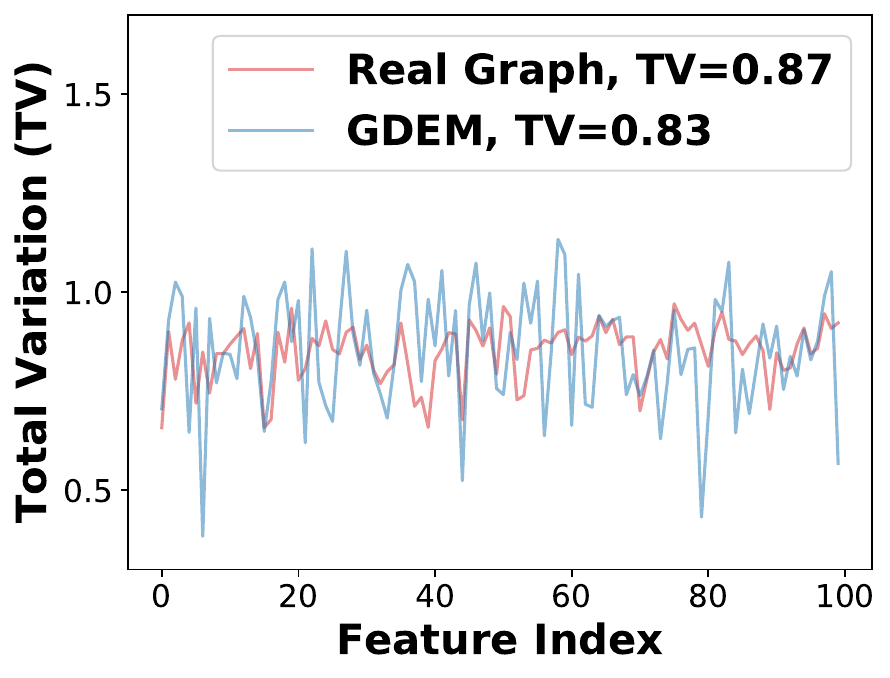}
    }
    \caption{TVs of synthetic graphs distilled by different methods.}
    \label{fig: visualize}
\end{figure}

We visualize the data distribution of synthetic graphs for a better understanding of our model.
Specifically, Figure~\ref{fig: visualize} illustrates the synthetic graphs distilled by GCOND, SGDD, and GDEM, from which we can observe that the value of TV in GDEM is the closest to the real graph.
SGDD is closer to the distribution of the real graph than GCOND, implying that SGDD can better preserve the structural information.
However, the performance is still not as good as GDEM, which validates the effectiveness of eigenbasis matching.

\begin{figure}[t]
    \centering
    \subfigure[Epoch: 50]{
        \centering
        \includegraphics[width=0.3\linewidth]{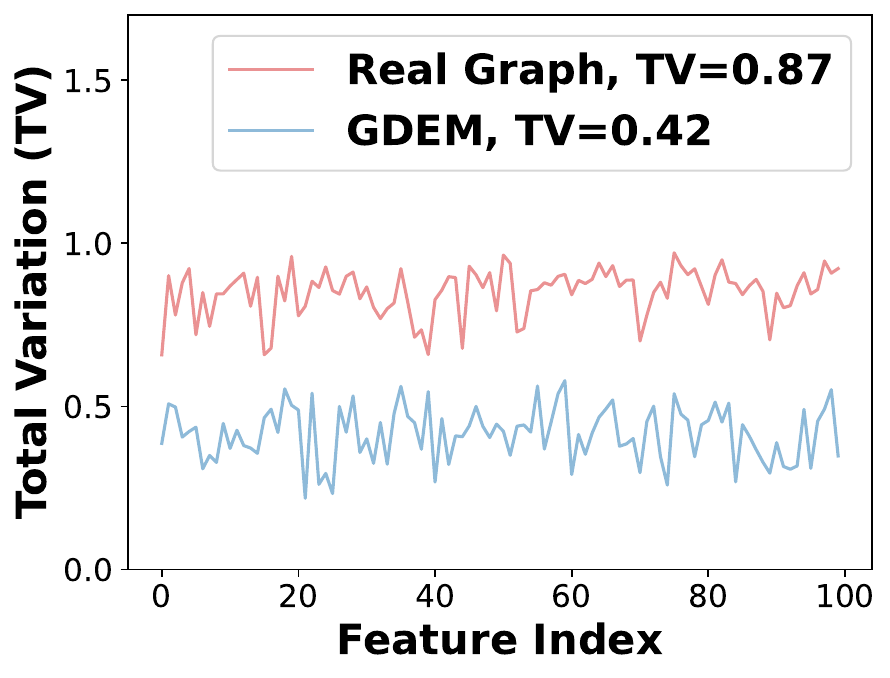}
    }
    \subfigure[Epoch: 100]{
        \centering
        \includegraphics[width=0.3\linewidth]{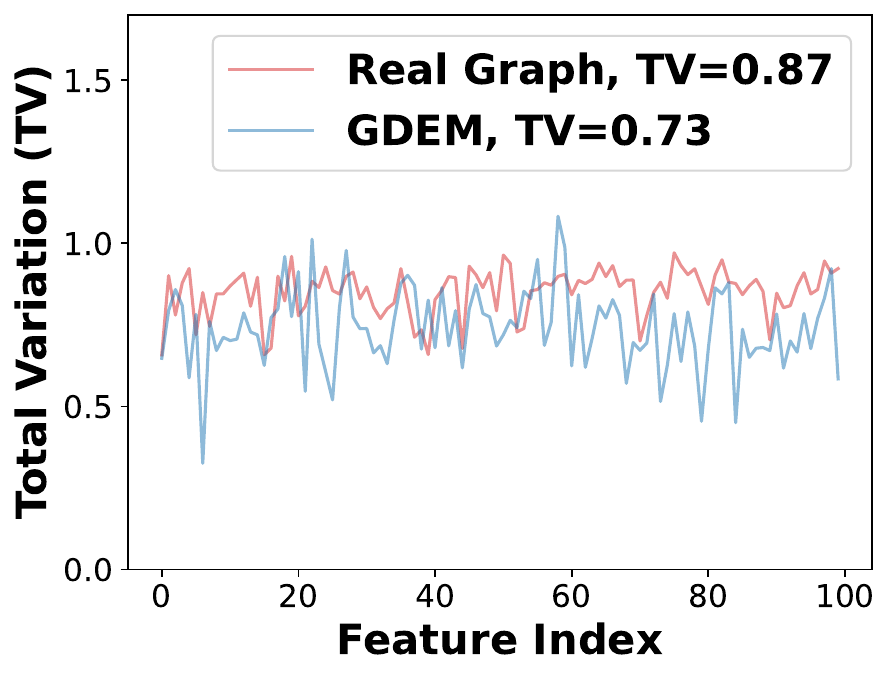}
    }
    \subfigure[Epoch: 200]{
        \centering
        \includegraphics[width=0.3\linewidth]{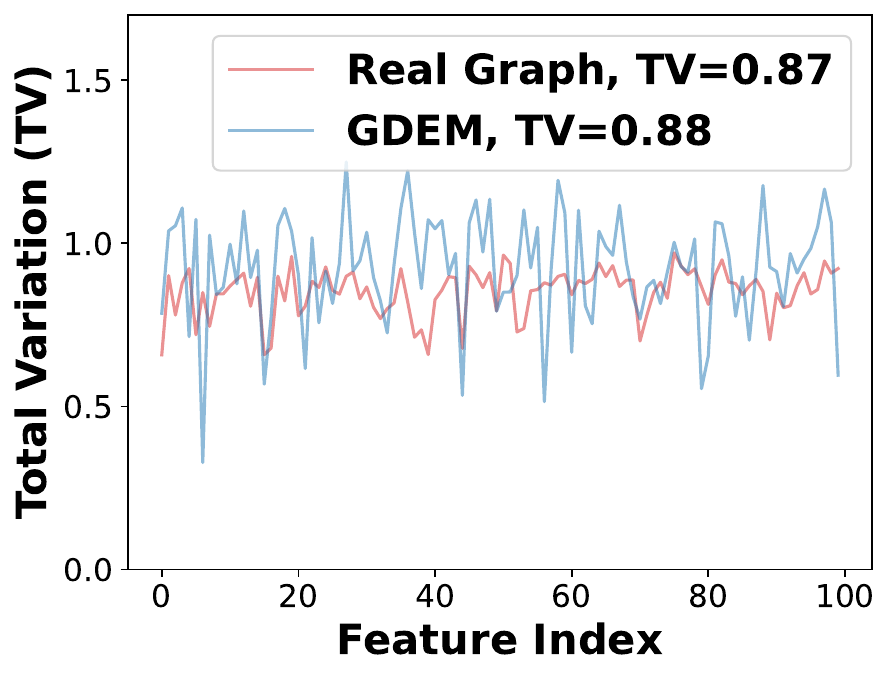}
    }
    \caption{TVs of synthetic graphs at different epochs (GDEM).}
    \label{fig: epoch}
\end{figure}

Besides, we also visualize the synthetic graphs distilled by GDEM at different epochs in Figure~\ref{fig: epoch}.
We can find that with the optimization of GDEM, the value of TV in the synthetic graphs is approaching the real graph (0.42 $\rightarrow$ 0.73 $\rightarrow$ 0.88), which validates Proposition~\ref{prop} that GDEM can preserve the spectral similarity of the real graph.

% In Figure~\ref{fig: TV_cora}, we visualize how the total variation of the synthetic graph evolves during the training process on the Cora dataset.
% Specifically, Figure~\ref{fig: TV0} illustrates the signal distribution at the first epoch, which has a lower classification accuracy due to random initialization.
% An interesting phenomenon emerges in Figures~\ref{fig: TV100} and \ref{fig: TV150}, where the total variations of the synthetic graphs at epochs 150 and 200 become smaller than the random initialization. Simultaneously, their performance notably improves.
% We guess this is because GDEM will first align the low-frequency subspaces of the real and synthetic graphs, which reduces the total variation and improves the model performance.
% Finally, we find that the synthetic graph at Epoch 200 demonstrates the best performance. It not only has the highest classification accuracy but also closely matches the signal distributions in the real graphs.
% This experiment validates Proposition~\ref{prop} that optimizing the objective of GDEM will reduce the difference in total variation.

\section{Related Work}

\noindent \textbf{Graph Neural Networks} aim to design effective convolution operators to exploit the node features and topology structure information adequately. GNNs have achieved great success in graph learning and play a vital role in diverse real-world applications~\cite{app3, yang2017neural}.
Existing methods are roughly divided into spatial and spectral approaches.
Spatial GNNs focus on neighbor aggregation strategies in the vertical domain~\cite{GCN, GAT, SAGE}.
Spectral GNNs aim to design filters in the spectral domain to extract certain frequencies for the downstream tasks~\cite{GPR-GNN, ChebNet, Specformer, ChebNetII, BernNet}.

\noindent \textbf{Dataset Distillation} (DD) has shown great potential in reducing data redundancy and accelerating model training~\cite{dd_survey1, dd_survey2, dd_survey3, dd_survey4}.
DD aims to generate small yet informative synthetic training data by matching the model gradient~\cite{DCGM, HaBa}, data distribution~\cite{DCDM, CAFE}, and training trajectory~\cite{DDMTT, guo2023towards} between the real and synthetic data.
As a result, models trained on the real and synthetic data will have comparable performance.

DD has been widely used for graph data, including node-level tasks, \eg, GCond~\cite{GCond}, SFGC~\cite{SFGC}, GCDM~\cite{GCDM} and MCond~\cite{MCond}, and graph-level tasks, \eg, DosCond~\cite{DosCond} and KIDD~\cite{KIDD}.
GCond is the first GD method based on gradient matching, which needs to optimize GNNs during the distillation procedure, resulting in inefficient computation. DosCond further provides one-step gradient matching to approximate gradient matching, thereby avoiding the bi-level optimization.
GCDM proposes distribution matching for GD, which views the receptive fields of a graph as its distribution. 
Additionally, SFGC proposes structure-free GD to compress the structural information into the node features.
KIDD utilizes the kernel ridge regression to further reduce the computational cost.
However, all these methods do not consider the influence of GNNs, resulting in spectrum bias and traversal requirement.

\section{Conclusion}
In this paper, we propose eigenbasis matching for graph distillation, which only aligns the eigenbasis and node features of the real and synthetic graphs, thereby alleviating the spectrum bias and traversal requirement of the previous methods.
Theoretically, GDEM preserves the restricted spectral similarity of the real graphs.
Extensive experiments on both homophilic and heterophilic graphs validate the effectiveness, generalization, and efficiency of the proposed method.
A promising future work is to explore eigenbasis matching without the need for explicit eigenvalue decomposition.

\section*{Acknowledgements}
This work is supported in part by the National Natural Science Foundation of China (No. U20B2045, 62192784, U22B2038, 62002029, 62172052).

\section*{Impact Statement}
This paper presents work aiming to advance the field of efficient graph learning and will save social resources by diminishing computation and storage energy consumption. There are many potential societal
consequences of our work, none of which we feel must be specifically highlighted here.

\bibliography{ref.bib}
\bibliographystyle{icml2024}

%%%%%%%%%%%%%%%%%%%%%%%%%%%%%%%%%%%%%%%%%%%%%%%%%%%%%%%%%%%%%%%%%%%%%%%%%%%%%%%
%%%%%%%%%%%%%%%%%%%%%%%%%%%%%%%%%%%%%%%%%%%%%%%%%%%%%%%%%%%%%%%%%%%%%%%%%%%%%%%
% APPENDIX
%%%%%%%%%%%%%%%%%%%%%%%%%%%%%%%%%%%%%%%%%%%%%%%%%%%%%%%%%%%%%%%%%%%%%%%%%%%%%%%
%%%%%%%%%%%%%%%%%%%%%%%%%%%%%%%%%%%%%%%%%%%%%%%%%%%%%%%%%%%%%%%%%%%%%%%%%%%%%%%
\newpage
\appendix
\onecolumn

\section{Experimental Details}

\subsection{Visualization of Synthetic Graphs}
\label{app: visualization}
\paragraph{Distillation Details with Low-pass and High-pass Filters.}
We use GCOND to distill two synthetic graphs on Pubmed by replacing SGC with a low-pass filter $\mathcal{F}_{L}=\mathbf{A}\mathbf{X}\mathbf{W}$ and a high-pass filter $\mathcal{F}_{H}=\mathbf{L}\mathbf{X}\mathbf{W}$, respectively.
% \begin{equation}
% \begin{aligned}
%     \mathcal{F}_{L}&=\mathbf{L}\mathbf{X}\mathbf{W}, \\
%     \mathcal{F}_{H}&=(\mathbf{I}-\mathbf{L})\mathbf{X}\mathbf{W}.
% \end{aligned}    
% \end{equation}

\paragraph{Visualization Details}
Once we generate the synthetic graphs, we calculate the value of total variation (TV) for each dimension. TV is a widely used metric to represent the distribution, i.e., smoothness, of a signal on the graph:
\begin{equation}
    \mathbf{x}^{\top} \mathbf{L} \mathbf{x} = \sum\nolimits_{(i, j) \in \mathcal{E}}(x_{i} - x_{j})^{2} = \sum_{i=1}^{n} \lambda_{i} \mathbf{x}^{\top} \mathbf{u}_{i} \mathbf{u}_{i}^{\top} \mathbf{x}.
\label{eq: total variation}
\end{equation}
Note that the edge number of synthetic graphs and the original graph is different, so we normalize node features and laplacian matrix first:
\begin{equation}
   \begin{aligned}
   {\hat{\mathbf{x}}}_i &= \frac{\mathbf{x}_i}{\left \| \mathbf{x}_i \right \|}, \\
   \hat{\mathbf{L}} &= \mathbf{I}_{N} - \mathbf{D}^{-\frac{1}{2}} \mathbf{A} \mathbf{D}^{-\frac{1}{2}},
\end{aligned} 
\end{equation}
where $\mathbf{x}_i$ is the $i$-th dimension node feature. Then we substitute $\hat{\mathbf{x}}_i$ and $\hat{\mathbf{L}}$ into Equation~\ref{eq: total variation} calculating the TV of the graph.\\
Additionally, we report the average TV of all dimensions as reported in the legend of the visualization figures.

\subsection{Cross-architecture Performance of GCOND and SGDD}
\label{app: cross}
To verify the cross-architecture performance of the GCOND and SGDD, we generate six synthetic graphs on Pubmed under a $0.15\%$ compression ratio, using six GNNs for the distillation procedure. Then we train these GNNs on the six synthetic graphs and evaluate their performance. Experimental settings are as follows.
\paragraph{Distillation Step.} For spatial GNNs, \ie, GCN, SGC, and PPNP, we set the aggregation layers to 2. For GCN, we use 256 hidden units for each convolutional layer. For spectral GNNs, \ie, ChebyNet, BernNet, and GPR-GNN, we set the polynomial order to 10. The linear feature transformation layers of all GNNs are set to 1. For hyper-parameters tuning, we select training epochs from \{400, 500, 600\}, learning rates of node feature and topology structure from \{0.0001, 0.0005, 0.001, 0.005, 0.05\}, outer loop from \{25, 20, 15, 10\}, and inner loop from \{15, 10, 5, 1\}.
\paragraph{Evaluation Step.} For spatial GNNs, we use two aggregation layers. For spectral GNNs, we set the polynomial order to 10. The hidden units of convolutional layers and linear feature transformation layers are both set to 256. We train each GNN for 2000 epochs and select the model parameters with the best performance on validation sets for evaluation.

\begin{table}[h]
    \centering
    \begin{minipage}{0.45\linewidth}
        \centering
        \caption{GCOND with various distillation ($\mathtt{D}$) and evaluation ($\mathtt{E}$) GNNs in Pubmed dataset.}
        \resizebox{\linewidth}{!}{
        \begin{tabular}{lcccccc}
        \toprule
        $\mathtt{D}$ $\mathlarger{\diagdown}$ $\mathtt{E}$ & GCN & SGC & \multicolumn{1}{c}{PPNP} & Cheb. & Bern. & GPR. \\
        \midrule
        GCN   & 74.57 & 71.70 & 75.53 & 70.13 & 68.40 & 71.73 \\
        SGC   & \textbf{77.72} & \textbf{77.60} & 77.34 & 76.03 & 74.42 & 76.52 \\
        PPNP  & 72.70 & 70.40 & 77.46 & 73.38 & 70.56 & 74.02 \\
        Cheb. & 73.60 & 70.62 & 75.10 & \textbf{77.30} & 77.62 & 78.10 \\
        Bern. & 67.68 & 73.76 & 74.30 & 77.20 & \textbf{78.12} & \textbf{78.28} \\
        GPR.  & 76.04 & 72.20 & \textbf{77.94} & 75.92 & 77.12 & 77.96 \\
        \midrule
        Optimal & 77.72 & 77.60 & 77.94 & 77.30 & 78.12 & 78.28 \\
        \bottomrule
        \end{tabular}}
        % \label{tab: pubmed_gcond}
    \end{minipage}
    \quad
    \begin{minipage}{0.45\linewidth}
        \centering
        \caption{SGDD with various distillation ($\mathtt{D}$) and evaluation ($\mathtt{E}$) GNNs in Pubmed dataset.}
        \resizebox{\linewidth}{!}{
        \begin{tabular}{lcccccc}
        \toprule
        $\mathtt{D}$ $\mathlarger{\diagdown}$ $\mathtt{E}$ & GCN & SGC & \multicolumn{1}{c}{PPNP} & Cheb. & Bern. & GPR. \\
        \midrule
        GCN   & 76.92 & 70.10 & 74.64 & 74.98 & 76.66 & 75.18 \\
        SGC   & \textbf{78.04} & \textbf{76.60} & \textbf{78.72} & 76.90 & 75.45 & 77.02 \\
        PPNP  & 76.44 & 74.34 & 76.28 & 73.70 & 74.94 & 75.98 \\
        Cheb. & 77.42 & 73.66 & 75.40 & \textbf{77.50} & \textbf{77.96} & 77.12 \\
        Bern. & 70.64 & 71.22 & 74.88 & 76.38 & 76.16 & 77.84 \\
        GPR.  & 63.76 & 61.24 & 76.32 & 71.40 & 71.70 & \textbf{78.30} \\
        \midrule 
        Optimal & 78.04 & 76.60 & 78.72 & 77.50 & 77.96 & 78.30 \\
        \bottomrule
        \end{tabular}}
        % \label{tab: pubmed_sgdd}
    \end{minipage}
\end{table}

\subsection{Implementation Details of GDEM}
\label{app: implementation}

% \noindent \textbf{Pre-processing of real graph.}
% Since the number of eigenvalues in the real graph is significantly greater than that in the synthetic graph,
% we can only match a few important eigenvectors.
% Specifically, we aim to choose the eigenvectors with the $K_1$ smallest and the $K_2$ largest eigenvalues, where $K_1$ and $K_2$ are hyperparameters, and $K_1 + K_2 \leq n^\prime$.
% This approach has been used in the graph coarsening algorithms~\cite{k1+k2}, as they encode the important frequency information of the graphs.
% The selected eigenvectors form a new eigenbasis, which is denoted as $\mathbf{U}_{K} = [\mathbf{u}_1, \cdots, \mathbf{u}_{K_1}, \mathbf{u}_{n-K_2}, \cdots, \mathbf{u}_{n}] \in \mathbb{R}^{n \times K}$.

\paragraph{Predefined Labels $\mathbf{Y}^\prime$ of Synthetic Graphs.}
The labels $\mathbf{Y}^\prime$ are predefined one-hot vectors, indicating the category to which the nodes belong. Specifically, given $N_l$ labeled nodes in the real graph, we set the number of nodes of category $c$ in the synthetic graph as $N_c^\prime=N_c \times \frac{N^\prime}{N_l}$, where $N_c$ is the number of nodes with label $c$. The setting will make the label distribution of the synthetic graph consistent with the real graph.

\paragraph{Initialization of Synthetic Graphs.}
Different from previous GD methods that directly learn the adjacency matrix of the synthetic graph, GDEM aims to generate its eigenbasis.
To ensure that the initialized eigenbasis is valid, we first use the stochastic block model (SBM) to randomly generate the adjacency matrix of the synthetic graph $\mathbf{A}^{\prime} \in \{0, 1\}^{N^\prime \times N^\prime}$, and then decompose it to produce the top-$K$ eigenvectors as the initialized eigenbasis
$\mathbf{U}_{K}^{\prime} \in \mathbb{R}^{N^\prime \times K}$.
Moreover, to initialize the synthetic node features $\mathbf{X}^{\prime} \in \mathbb{R}^{N^{\prime} \times d}$, we first train an MLP $\rho(\cdot)$ in the real node features. Then we freeze the well-trained MLP and feed the synthetic node features into it to minimize the classification objective. This process can be formulated as:
\begin{equation}
    \min_{\mathbf{X}^{\prime}} \sum_{i=1}^{n^\prime} - y_{i}^{\prime} \log \rho(\mathbf{x}_{i}^{\prime}, \theta^*), \text { s.t. } \theta^{*} = \underset{\theta}{\arg \min} \sum_{i=1}^{n} - y_{i} \log \rho(\mathbf{x}_{i}, \theta)
\label{eq: initialization}
\end{equation}
where $\theta$ indicates the parameters of MLP.

\subsection{Complexity of Different Methods}
\label{app: complexity}
We analyze the complexity of different methods and give the final complexity in Table~\ref{tab: complexity}. We use $E$ to present the number of edges. For simplicity, we use $d$ to denote both feature dimension and hidden units of GNNs. $t$ is the number of GNN layers and $r$ is the number of sampled neighbors per node. $\theta_t$ denotes the model parameters of the GNNs. For SFGC, $M$ is the number of training trajectories and $S$ is the length of each trajectory.

\paragraph{Complexity of GDEM.}\quad

(1) Pre-processing: The complexity of decomposition is $\mathcal{O}(KN^{2})$. It’s noteworthy that the decomposition is performed once per graph and can be repeatedly used for subsequent training, inference, and hyperparameter tuning. Therefore, the time overhead of decomposition should be amortized by the entire experiment rather than simply summarized them. Additionally, we pre-process ${\mathbf{u}_{k}^{\top} \mathbf{X}}$ in Equation~\ref{eq: le} and $H$ in Equation~\ref{eq: ld}, which cost $\mathcal{O}(KNd)$ and $\mathcal{O}(Ed)$. \\
(1) Complexity of $\mathcal{L}_{e}$: $\mathcal{O}(KN^{\prime}d+Kd^{2})$. \\
(2) Complexity of $\mathcal{L}_{d}$: The complexity of calculating $H'$ is $\mathcal{O}(KN^{\prime}d^{\prime})$. The calculation of cosine similarity costs $\mathcal{O}(Cd^2)$. \\
(3) Complexity of $\mathcal{L}_{o}$: $\mathcal{O}(KN^{\prime2})$. \\
The final complexity can be simplified as $\mathcal{O}(KN^2+KNd+Ed)+\mathcal{O}(KN^{\prime2}+KN^{\prime}d+(K+C)d^2)$.

\paragraph{Complexity of GCOND.}\quad

(1) Pre-processing: GCOND doesn’t need special pre-processing.\\
(2) Inference for $A^{\prime}$: $\mathcal{O}(N^{\prime2}d^{2})$.\\
(3) Forward process of SGC on the original graph: $\mathcal{O}(r^{t}Nd^{2})$. That on the synthetic graph: $\mathcal{O}(tN^{\prime2}d+tN^{\prime}d)$.\\
(4) Calculation of second-order derivatives in backward propagation: $\mathcal{O}(|\theta_t|+|A^{\prime}|+|X^{\prime}|)$.\\
The final complexity can be simplified as $\mathcal{O}(r^{t}Nd^{2})+\mathcal{O}(N^{\prime2}d^{2})$.

\paragraph{Complexity of SGDD.} \quad

(1) Pre-processing: SGDD doesn’t need special pre-processing.\\
(2) Inference for $A^{\prime}$: $\mathcal{O}(N^{\prime2}d^{2})$.\\
(3) Forward process of SGC on the original graph: $\mathcal{O}(r^{t}Nd^{2})$. That on the synthetic graph: $\mathcal{O}(tN^{\prime2}d+tN^{\prime}d)$.\\
(4) Calculation of second-order derivatives in backward propagation: $\mathcal{O}(|\theta_t|+|A^{\prime}|+|X^{\prime}|)$.\\
(5) Structure optimization term: $\mathcal{O}(N^{\prime2}k+NN^{\prime2})$. \\
The final complexity can be simplified as $\mathcal{O}(r^{t}Nd^{2})+\mathcal{O}(N^{\prime2}N)$.

\paragraph{Complexity of SFGC.} \quad

(1) Pre-processing: $\mathcal{O}(MS(tEd+tNd^{2}))$. Note that $MS$ is usually very large, so it cannot be omitted.\\
(2) Forward process of GCN on the synthetic graph: $\mathcal{O}(tN^{\prime}d^{2}+tN^{\prime}d)$. Note that SFGC pre-trains the trajectories on GCN, so there is no need to calculate the forward process on the original graph.\\
(3) Backward propagation: SFGC uses a MTT\cite{DDMTT} method, which results in bi-level optimization\cite{dd_survey4} for the backward.\\
The final complexity can be simplified as $\mathcal{O}(MS(tEd+tNd^{2}))+\mathcal{O}(tN^{\prime}d^{2})$.
% \begin{table}[h]
%   \centering
%   \caption{Complexity of different distillation methods.}
%   % \resizebox{0.5\linewidth}{!}{
%     \begin{tabular}{lcl}
%     \toprule
%     Method & Pre-processing & Training \\
%     \midrule
%     GCOND & - & $\mathcal{O}(r^{L}Nd^{2})+\mathcal{O}(N^{\prime2}d^{2})$ \\
%     SGDD  & - & $\mathcal{O}(r^{L}Nd^{2})+\mathcal{O}(N^{\prime2}N)$ \\
%     SFGC  & $\mathcal{O}(MS(LEd+LNd^{2}))$ & $\mathcal{O}(LN^{\prime}d^{2})$\\
%     GDEM  & $\mathcal{O}(KN^{2})$ & $\mathcal{O}((K+d)N^{2} + (K+C)Nd)$ \\
%     \bottomrule
%     \end{tabular}
%   \label{tab: complexity}%
% \end{table}%

\begin{table}[h]
  \centering
  \caption{Complexity of different distillation methods.}
  % \resizebox{0.5\linewidth}{!}{
    \begin{tabular}{lcc}
    \toprule
    Method & Pre-processing & Training \\
    \midrule
    GCOND & - & $\mathcal{O}(r^{L}Nd^{2})+\mathcal{O}(N^{\prime2}d^{2})$ \\
    SGDD  & - & $\mathcal{O}(r^{L}Nd^{2})+\mathcal{O}(N^{\prime2}N)$ \\
    SFGC  & $\mathcal{O}(MS(LEd+LNd^{2}))$ & $\mathcal{O}(LN^{\prime}d^{2})$\\
    GDEM  & $\mathcal{O}(KN^2+KNd+Ed)$ & $\mathcal{O}(KN^{\prime2}+KN^{\prime}d+(K+C)d^2)$ \\
    \bottomrule
    \end{tabular}
  \label{tab: complexity}%
\end{table}%

\subsection{Statistics of Datasets}
\label{app: statistics}

In the experiments, we use seven graph datasets to validate the effectiveness of GDEM. For homophilic graphs, we use the public data splits. For heterophilic graphs, we use the splitting with training/validation/test sets accounting for 2.5\%/2.5/\%95\% on Squirrel, and 50\%/25\%25\% on Gamers. The detailed statistical information of each dataset is shown in Table~\ref{tab: dataset details}.

\begin{table}[h]
  \centering
  \caption{Statistics of datasets.}
  \resizebox{0.9\linewidth}{!}{
    \begin{tabular}{lrrrrrrr}
    \toprule
    \textbf{Dataset} & \textbf{Nodes} & \textbf{Edges} & \textbf{Classes} & \textbf{Features} & \textbf{Training/Validation/Test} & \textbf{Edge hom.} & \textbf{LCC}\\
    \midrule
    Citeseer & 3,327 & 4,732 & 6 & 3,703 & 120/500/1000 & 0.74 & 2,120\\
    Pubmed & 19,717 & 44,338 & 3 & 500 & 60/500/1,000 & 0.80 & 19,717\\
    Ogbn-arxiv & 169,343 & 1,166,243 & 40 & 128 & 90,941/29,799/48,603 & 0.66 & 169,343 \\
    Flickr & 89,250 & 899,756 & 7 & 500 & 44,625/22,312/22,313 & 0.33 & 89,250 \\
    Reddit & 232,965 & 57,307,946 & 41 & 602 & 153,932/23,699/55,334 & 0.78 & 231,371 \\
    Squirrel & 5,201 & 396,846 & 5 & 2,089 & 130/130/4,941 & 0.22 & 5,201\\
    Gamers & 168,114 & 13,595,114 & 2 & 7 & 84,056/42,028/42,030 & 0.55 & 168,114\\
    \bottomrule
    \end{tabular}}
  \label{tab: dataset details}
\end{table}

\subsection{Baselines}
\label{app: baseline}
For a fair comparison of performance, we adopt the results of baselines reported in their papers, which are evaluated through meticulous experimental design and careful hyperparameter tuning. The experimental details are as follows:
(1) GCOND employs a 2-layer SGC for distillation and a 2-layer GCN with 256 hidden units for evaluation. \\
(2) SGDD employs a 2-layer SGC for distillation and a 2-layer GCN with 256 hidden units for evaluation. \\
(3) SFGC employs 2-layer GCNs with 256 hidden units both for distillation and evaluation.

\subsection{Evaluation Details}
\label{app: evaluation}
\paragraph{Performance Evaluation.}
For comparison with baselines, we report the performance of GDEM evaluated with a 2-layer GCN with 256 hidden units. Specifically, we generate 10 synthetic graphs with different seeds on the original graph. Then we train the GCN using these 10 synthetic graphs and report the average results of the best performance evaluated on test sets of the original graph.
\paragraph{Generalization Evaluation.}
For generalization evaluation, we train 6 GNNs using the synthetic graphs generated by different distillation methods. For SGC, GCN, and APPNP, we use 2-layer aggregations. For ChebyNet, we set the convolution layers to 2 with propagation steps from \{2, 3, 5\}.
For BernNet and GPRGNN, we set the polynomial order to 10. The hidden units of both convolution layers and linear feature transformation are 256.

\subsection{Hyperparamters}
\label{app: hyperparameter}
Hyperparameter details are listed in Table~\ref{tab: hyperparameters}. $\tau_1$ and $\tau_2$ are steps for alternating updates of node features and eigenvectors. $\alpha$, $\beta$, and $\gamma$ denote the weights in Equation~\ref{eq: l_total}. lr\_feat and lr\_eigenvecs are the learning rates of node features and eigenvectors, respectively.

\begin{table}[h]
  \centering
  \caption{Hyper-parameters of GDEM.}
    \begin{tabular}{lccccccccccc}
    \toprule
    \multicolumn{1}{l}{Dataset} & \multicolumn{1}{l}{Ratio} & epochs & $K_1$ & $K_2$ & $\tau_1$ & $\tau_2$ & $\alpha$ & $\beta$ & $\gamma$ & lr\_feat & lr\_eigenvecs \\
    \midrule
    \multirow{3}[0]{*}{Citeseer} & 0.90\% & 500 & 30 & 0 & 5 & 1 & 1.0 & 1e-05 & 1.0 & 0.0001 & 0.01 \\
      & 1.80\% & 1500 & 48 & 12 & 10 & 15 & 0.05 & 1e-05 & 0.5 & 0.0005 & 0.0005 \\
      & 3.60\% & 500 & 114 & 6 & 1 & 10 & 0.01 & 1e-06 & 0.1 & 0.001 & 0.0001 \\
      \midrule
    \multirow{3}[0]{*}{Pubmed} & 0.08\% & 1000 & 15 & 0 & 15 & 5 & 0.0001 & 1e-07 & 0.01 & 0.0001 & 0.0005 \\
      & 0.15\% & 1500 & 30 & 0 & 5 & 5 & 1.0 & 1e-05 & 0.01 & 0.0005 & 0.01 \\
      & 0.30\% & 1500 & 57 & 3 & 20 & 1 & 0.01 & 1e-07 & 0.5 & 0.001 & 0.0001 \\
  \midrule
    \multirow{3}[0]{*}{Ogbn-arxiv} & 0.05\% & 500 & 86 & 4 & 1 & 5 & 0.0001 & 1e-02 & 0.01 & 0.0005 & 0.0005 \\
      & 0.25\% & 2000 & 409 & 45 & 10 & 5 & 0.01 & 1e-04 & 0.01 & 0.0001 & 0.0001 \\
      & 0.50\% & 1000 & 773 & 136 & 1 & 5 & 0.001 & 1e-04 & 1.0 & 0.0001 & 0.005 \\
  \midrule
    \multirow{3}[0]{*}{Flickr} & 0.10\% & 2000 & 44 & 0 & 5 & 10 & 0.01 & 1e-07 & 0.05 & 0.0001 & 0.05 \\
      & 0.50\% & 2000 & 223 & 0 & 5 & 10 & 0.01 & 1e-07 & 0.05 & 0.0001 & 0.05 \\
      & 1.00\% & 2000 & 446 & 0 & 5 & 10 & 0.01 & 1e-07 & 0.05 & 0.0001 & 0.05 \\
  \midrule
    \multirow{3}[0]{*}{Reddit} & 0.05\% & 1000 & 76 & 0 & 20 & 5 & 1.0 & 1e-06 & 0.01 & 0.0001 & 0.0001 \\
      & 0.10\% & 500 & 153 & 0 & 15 & 10 & 0.5 & 1e-06 & 05 & 0.0005 & 0.005 \\
      & 0.50\% & 1000 & 693 & 76 & 5 & 5 & 1.0 & 1e-06 & 0.5 & 0.0005 & 0.0001 \\
  \midrule
    \multirow{3}[0]{*}{Squirrel} & 0.60\% & 1000 & 31 & 1 & 5 & 1 & 1.0 & 1e-07 & 0.01 & 0.0001 & 0.005 \\
      & 1.20\% & 500 & 62 & 3 & 10 & 5 & 1.0 & 1e-07 & 0.01 & 0.0001 & 0.0001 \\
      & 2.05\% & 2000 & 104 & 26 & 5 & 1 & 0.0001 & 1e-05 & 0.05 & 0.0001 & 0.01 \\
  \midrule
    \multirow{3}[0]{*}{Gamers} & 0.05\% & 2000 & 80 & 4 & 15 & 1 & 0.0001 & 1e-07 & 0.05 & 0.0001 & 0.01 \\
      & 0.25\% & 2000 & 420 & 0 & 20 & 20 & 0.0001 & 1e-07 & 0.05 & 0.0001 & 0.005 \\
      & 0.50\% & 500 & 756 & 84 & 15 & 1 & 0.0001 & 1e-07 & 0.05 & 0.0001 & 0.0001 \\
    \bottomrule
    \end{tabular}%
  \label{tab: hyperparameters}%
\end{table}%

% \subsection{Full results of the generalization experiment}
% \label{app: cross_full}

% Table~\ref{tab: generalization_app} shows the generalization ability of baselines on Citeseer, Pubmed, and Squirrel.

\subsection{Analysis of the Worse Performance on Obgn-arxiv}
\label{app: analysis of arXiv}
To investigate the reason why GDEM performs slightly worse on Obgn-arxiv but achieves promising results on other large-scale graphs, we evaluate the number of useful eigenbasis in both Ogbn-arxiv and Reddit. Specifically, we first truncate the graph structures of Ogbn-arxiv and Reddit by:
\begin{equation}
    \mathbf{A}^\prime=\sum_{k=1}^{K_1} \lambda_k \mathbf{u}_{k} \mathbf{u}_{k}^{\top} + \sum_{k=N-K_2+1}^{N} \lambda_k \mathbf{u}_{k} \mathbf{u}_{k}^{\top}
\end{equation}
where $K_1=r_kK$ and $K_2=(1-r_k)K$. We then gradually increase the value of $K$ and train a 2-layer SGC on each truncated graph structure. The results are shown in Table~\ref{tab: useful eigenbasis}.
\begin{table*}[t]
  \centering
  \caption{The node classification performance of Ogbn-arxiv and Reddit on various truncated graph structures.}
  \begin{tabular}{cccccc}
  \toprule
    Dataset & $K=500$ & $K=1000$ & $K=3000$ & $K=5000$ & Full Graph \\
    \midrule
    Reddit & 92.41±0.49 & 93.45±0.48 & 93.94±0.41 & 94.07±0.37 & 94.51±0.24 \\
    \midrule
    Ogbn-arxiv & 61.87±0.89 & 64.65±1.20 & 67.32±1.11 & 69.22±0.93 & 70.02±1.19 \\
    \bottomrule
    \end{tabular}%
  \label{tab: useful eigenbasis}%
\end{table*}%

We can observe that in Reddit, only 1,000 eigenvectors are enough to match the performance of the full graph (93.45 / 94.51 $\approx$ 98.9\%), while in Ogbn-arxiv, a large number of eigenvectors (5,000) is required to approximate the full graph (69.22 / 70.02 $\approx$ 98.9\%). Thus, we speculate that the structure information of Ogbn-arxiv is more widely distributed in the eigenbasis, making it challenging for GDEM to compress the entire distribution in synthetic data with an extremely small compression rate.

\section{Theoretical Analysis of RSS for Gradient Matching}
We further theoretically analyze whether the gradient matching method can preserve the restricted spectral similarity. Given $x^\prime$ and $L^\prime$ learned by the gradient matching method, we have:

\begin{equation}
\begin{aligned}
 &\left|\mathbf{x}^{\top} \mathbf{L} \mathbf{x}-\mathbf{x}^{\prime^{\top}} \mathbf{L}^{\prime} \mathbf{x}^{\prime}\right| \\
 =&\left|\sum_{k=0}^K \lambda_k \mathbf{x}^{\top} \mathbf{u}_k \mathbf{u}_k^{\top} \mathbf{x} - \sum_{k=0}^K \lambda^\prime_k {\mathbf{x}^\prime}^{\top} \mathbf{u}_k^\prime {\mathbf{u}_k^\prime}^{\top} \mathbf{x}^\prime\right| \\
 =&\left|\left(\sum_{k=0}^K \lambda_k \mathbf{x}^{\top} \mathbf{u}_k \mathbf{u}_k^{\top} \mathbf{x} - \sum_{k=0}^K \lambda_k {\mathbf{x}^\prime}^{\top} \mathbf{u}_k^\prime {\mathbf{u}_k^\prime}^{\top} \mathbf{x}^\prime \right) +\left(\sum_{k=0}^K \lambda_k {\mathbf{x}^\prime}^{\top} \mathbf{u}_k^\prime {\mathbf{u}_k^\prime}^{\top} \mathbf{x}^\prime- \sum_{k=0}^K \lambda_k^\prime {\mathbf{x}^\prime}^{\top} \mathbf{u}_k^\prime {\mathbf{u}_k^\prime}^{\top} \mathbf{x}^\prime\right)\right| \\
 \leqslant &\sum_{k=0}^K\lambda_k\left|\mathbf{x}^{\top} \mathbf{u}_k \mathbf{u}_k^{\top} \mathbf{x}-{\mathbf{x}^\prime}^{\top} \mathbf{u}_k^\prime {\mathbf{u}_k^\prime}^{\top} \mathbf{x}^\prime \right|+\sum_{k=0}^K\left|\lambda_k-\lambda_k^\prime \right|\left({\mathbf{x}^\prime}^{\top} \mathbf{u}_k^\prime {\mathbf{u}_k^\prime}^{\top} \mathbf{x}^\prime \right)  
\end{aligned}
\end{equation}

Combining with Lemma~\ref{collpse_GCN}, when the number of GCN layers goes to infinity, the objective optimization based on gradient matching is dominated by $\left|\mathbf{x}^{\top} \mathbf{u}_0 \mathbf{u}_0^{\top} \mathbf{x}-{\mathbf{x}^\prime}^{\top} \mathbf{u}_0^\prime {\mathbf{u}_0^\prime}^{\top} \mathbf{x}^\prime \right|$, while paying less attention to the optimization of $\left|\mathbf{x}^{\top} \mathbf{u}_k \mathbf{u}_k^{\top} \mathbf{x}-{\mathbf{x}^\prime}^{\top} \mathbf{u}_k^\prime {\mathbf{u}_k^\prime}^{\top} \mathbf{x}^\prime \right|$, when $k\neq 0$. Thus, gradient matching fails to constrain the first term of the upper bound of RSS. Moreover, gradient matching introduces spectrum bias causing $\lambda_k^{\prime} \neq  \lambda_k$, thus failing to constrain the second term of the upper bound. In summary, the gradient matching method is unable to preserve the restricted spectral similarity.

\section{Graph Distiilation}
\label{app: gd}

\paragraph{Gradient Matching}~\cite{GCond, DosCond} generates the synthetic graph and node features by minimizing the differences between model gradients on $\mathcal{G}$ and $\mathcal{G}^{\prime}$, which can be formulated as:
\begin{equation}
    \min_{\mathbf{A}^{\prime}, \mathbf{X}^{\prime}} \underset{\theta \sim P_{\theta}}{\mathbb{E}} 
    \left[ D \left( \nabla_\theta \mathcal{L} \left( \Phi_\theta \left(\mathbf{A}^{\prime}, \mathbf{X}^{\prime}\right), \mathbf{Y}^{\prime}\right), \nabla_\theta \mathcal{L} \left( \Phi_\theta \left( \mathbf{A}, \mathbf{X} \right), \mathbf{Y} \right)\right)\right],
\end{equation}
where $\Phi_\theta$ is the condensation GNNs with parameters $\theta$, $\nabla_\theta$ indicates the model gradients, $D$ is a metric to measure their differences, and $\mathcal{L}$ is the loss function.
For clarity, we omit the subscript that indicates the training data.

\paragraph{Distribution Matching}~\cite{GCDM} aims to align the distributions of node representations in each GNN layer to generate the synthetic graph, which can be expressed as:
\begin{equation}
    \min_{\mathbf{A}^{\prime}, \mathbf{X}^{\prime}} \underset{\theta \sim P_{\theta}}{\mathbb{E}} 
    \left[\sum_{t=1}^{L} D \left(\Phi^{t}_{\theta} \left( \mathbf{A}^{\prime}, \mathbf{X}^{\prime} \right), \Phi^{t}_{\theta} \left( \mathbf{A}, \mathbf{X} \right) \right)\right],
\end{equation}
where $\Phi^{t}_{\theta}$ is the $t$-th layer in GNNs.

\paragraph{Trajectory Matching}~\cite{SFGC} aligns the long-term GNN learning behaviors between the original graph and the synthetic graph:
\begin{equation}
\min_{\mathbf{A}^{\prime}, \mathbf{X}^{\prime}} \underset{\theta_t^{\ast ,i} \sim P_{\Theta^{\mathcal{T}}}}{\mathbb{E}} \left [ \mathcal{L}_{\text{meta-tt}}\left ( \theta_{t}^{\ast }|_{t=t_0}^{p}, \tilde{\theta}_{t}|_{t=t_0}^{q} \right ) \right ].
\end{equation}
where $\theta_{t}^{\ast }|_{t=t_0}^{p}$ and $\tilde{\theta}_{t}|_{t=t_0}^{q}$ is the parameters of $\text{GNN}_{\mathcal{T}}$ and $\text{GNN}_{\mathcal{S}}$, $\mathcal{L}_\text{meta-tt}$ calculates certain parameter training intervals within $\left [ \theta_{t_0}^{\ast ,i},\theta_{t_0+p}^{\ast ,i} \right ]$ and $\left [ \tilde{\theta}_{t_0},\tilde{\theta}_{t_0+q} \right ]$.

\section{General Settings}
\paragraph{Optimizer.} We use the Adam optimizer for all experiments.
\paragraph{Environment.} The environment in which we run experiments is:
\begin{itemize}
\setlength{\itemsep}{0.7ex}
    \item Linux version: 5.15.0-91-generic
    \item Operating system: Ubuntu 22.04.3 LTS
    \item CPU information: Intel(R) Xeon(R) Platinum 8358 CPU @ 2.60GHz
    \item GPU information: NVIDIA A800 80GB PCIe
\end{itemize}
\paragraph{Resources.} The address and licenses of all datasets are as follows:
\begin{itemize}
\setlength{\itemsep}{0.7ex}
    \item Citeseer: https://github.com/kimiyoung/planetoid (MIT License)
    \item Pubmed: https://github.com/kimiyoung/planetoid (MIT License)
    \item Ogbn-arxiv: https://github.com/snap-stanford/ogb (MIT License)
    \item Flickr: https://github.com/GraphSAINT/GraphSAINT (MIT License)
    \item Reddit: https://github.com/williamleif/GraphSAGE (MIT License)
    \item Squirrel: https://github.com/benedekrozemberczki/MUSAE (GPL-3.0 license)
    \item Gamers: https://github.com/benedekrozemberczki/datasets (MIT License)
\end{itemize}

\end{document}